\documentclass[journal]{IEEEtran}
\ifCLASSINFOpdf
  \usepackage[pdftex]{graphicx}
  \usepackage[nocompress]{cite}
  \usepackage{graphicx}
  \usepackage{amsmath}
  \usepackage{amssymb}
  \usepackage{booktabs}
  \usepackage{algorithm}
  \usepackage{algorithmic}
  \usepackage{multirow}
  \usepackage[dvipsnames,table,xcdraw]{xcolor}
  \usepackage{tikz}
  \usepackage{booktabs}
  \usepackage{threeparttable}
  \usepackage{graphicx}                                                           
  \usepackage{float} 
  \usepackage{times}
  \usepackage{epsfig}
  \usepackage{subfigure}
\else
\fi
\begin{document}
%
\title{Unsupervised Gait Recognition with \\ Selective Fusion}

\author{Xuqian Ren,
        Shaopeng Yang,
        Saihui Hou,
        Chunshui Cao,
        Xu Liu and
        Yongzhen Huang
\thanks{Xuqian Ren is with the Computer Science Unit, Faculty of Information Technology and Communication Sciences, Tampere University, Tampere 33720, Finland. This work is finished when she was an intern at Watrix Technology Limited Co. Ltd before becoming a Ph.D. candidate at Tampere University.
(E-mail: xuqian.ren@tuni.fi)}
\thanks{Shaopeng Yang is with School of Artificial Intelligence,
Beijing Normal University, Beijing 100875, China and also with Watrix Technology Limited Co. Ltd, Beijing 100088, China. He is co-first author.}
\thanks{Saihui Hou and Yongzhen Huang are with School of Artificial Intelligence,
Beijing Normal University, Beijing 100875, China and also with Watrix Technology Limited Co. Ltd, Beijing 100088, China. (E-mail: housai hui@bnu.edu.cn, huangyongzhen@bnu.edu.cn). They are the corresponding authors of this paper.}
\thanks{Chunshui Cao and Xu Liu are with Watrix Technology Limited Co. Ltd, Beijing 100088, China.}
}

%
%

\markboth{Journal of \LaTeX\ Class Files,~Vol.~14, No.~8, August~2015}%
{Shell \MakeLowercase{\textit{et al.}}: Bare Demo of IEEEtran.cls for IEEE Journals}
%



\maketitle

\begin{abstract}
Previous gait recognition methods primarily relied on labeled datasets, which require a labor-intensive labeling process. To eliminate this dependency, we focus on a new task: Unsupervised Gait Recognition (UGR). We introduce a cluster-based baseline to solve UGR. However, we identify additional challenges in this task. First, sequences of the same person in different clothes tend to cluster separately due to significant appearance changes. Second, sequences captured from $0^{\circ}$ and $180^{\circ}$ views lack distinct walking postures and do not cluster with sequences from other views. To address these challenges, we propose a Selective Fusion method, consisting of Selective Cluster Fusion (SCF) and Selective Sample Fusion (SSF). SCF merges clusters of the same person wearing different clothes by updating the cluster-level memory bank using a multi-cluster update strategy. SSF gradually merges sequences taken from front/back views using curriculum learning. Extensive experiments demonstrate the effectiveness of our method in improving rank-1 accuracy under different clothing and view conditions.
\end{abstract}

\begin{IEEEkeywords}
Gait Recognition, Unsupervised Learning, Contrastive Learning, Curriculum Learning.
\end{IEEEkeywords}

%
\IEEEpeerreviewmaketitle

\section{Introduction}
\IEEEPARstart{W}{ith} the growing intelligent security and safety camera systems, gait recognition has gradually gained more attention and exploration for its non-contact, long-term, and long-distance recognition properties. 
Several works~\cite{chao2019gaitset,fan2020gaitpart,lin2021gait} attempt to solve gait recognition tasks and have reached significant progress in a laboratory environment. 
However, gait recognition in a realistic situation~\cite{ren2022progressive} and will be affected by many factors such as occlusion, dirty labels, labeling, and more. 
Labeling, especially, is a major challenge, requiring intensive manual effort for pairwise data. Therefore, training on unlabeled datasets becomes crucial to save resources and address these challenges.
%
%

\begin{figure}[t]
	\centering	 
	\includegraphics[width=\linewidth]{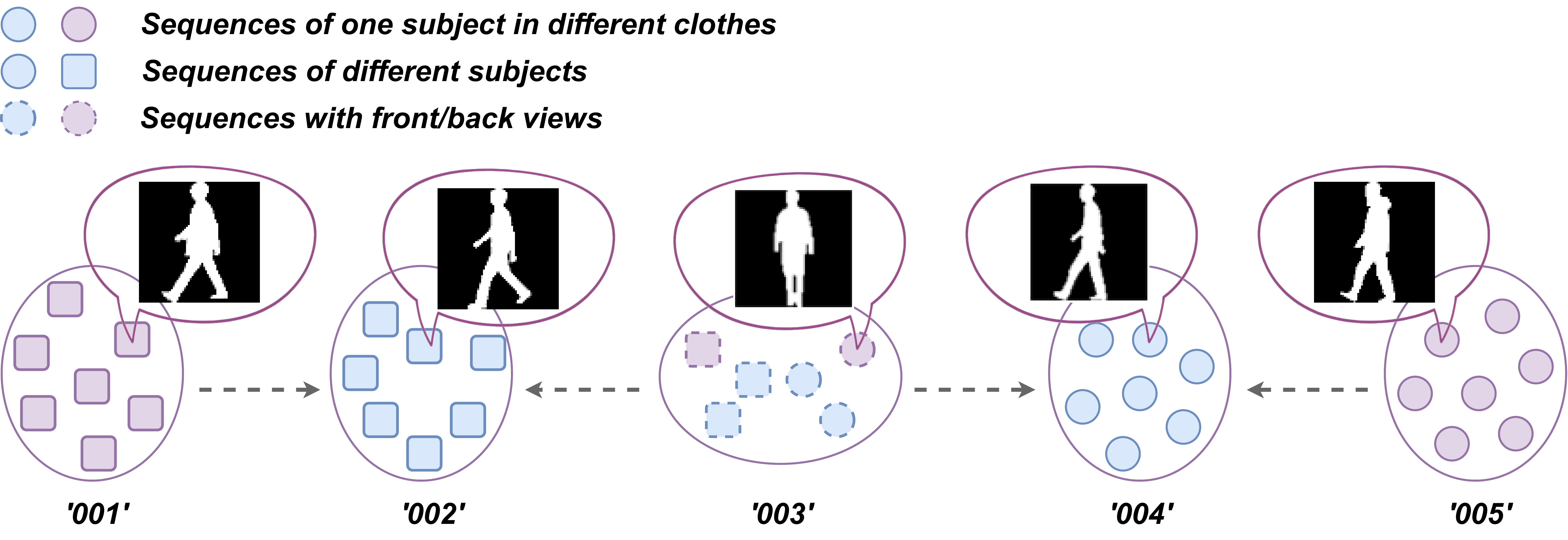}	 	
	\caption{Two main challenges in UGR. A kind of style in different colors denotes a subject in different clothes, which are usually erroneously assigned with different pseudo labels (\textit{e.g.}, `001', `002'). Also, sequences taken from front/back views of different subjects tend to mix together (\textit{e.g.}, `003').}
	\label{fig:Introduction}
\end{figure}

To realize gait recognition using an unlabeled dataset, we focus on a task called \textbf{Unsupervised Gait Recognition} (UGR) to facilitate the research on training gait recognition models with new unlabeled datasets.
Here we focus on using silhouettes for gait recognition to illustrate our method.
When only using silhouettes of human walking sequences as input, due to lack of enough information, we observe two main challenges in UGR, as shown in Figure~\ref{fig:Introduction}.
First, due to the large change in appearance, sequences in different clothes of a subject are hard to gather into one cluster without any label supervision.
Second, sequences captured from front/back views, such as views in $0^{\circ}/018^{\circ}/162^{\circ}/180^{\circ}$ in CASIA-B~\cite{yu2006framework}, are challenging to gather with sequences taken from other views of the same person because they lack vital information, such as walking postures.
Furthermore, these sequences tend to cluster into small groups based on their views or get mixed with sequences of the same perspective from other subjects.
So in this paper, we provide methods to overcome them accordingly.

Currently, some Person Re-identification (Re-ID) works~\cite{lin2019bottom,ge2020self,lin2020unsupervised,wang2020unsupervised,dai2021cluster,9978648} have already touched the field of identifying person in an unsupervised manner.
There are also some traditional methods, such as~\cite{ball2012unsupervised,cola2015unsupervised,rida2015unsupervised} employ unsupervised learning to facilitate the development of gait recognition. 
However, research directions involving methods based on deep learning in this field are still under-explored.
In this paper, we use the state-of-the-art pattern in~\cite{dai2021cluster}, a cluster-based framework with contrastive learning, as a baseline to realize UGR.
%
%
To address the two challenges, we propose a new method called \textbf{S}elective \textbf{F}usion(SF), to gradually pull cross-view and cross-cloth sequences together.

Our method comprises two techniques: Selective Cluster Fusion (SCF), which is used to narrow the distance of cross-cloth clusters, and Selective Sample Fusion (SSF), which is used to pull cross-view pairs nearly, especially helpful for outliers near $0^{\circ}/180^{\circ}$.
To be specific, first, in SCF, we use a support set selection module to generate a support set for each cluster.
In the support set, there are selected candidate clusters of each corresponding cluster that potentially belong to the same person but are in different clothes.
There is another multi-cluster update strategy designed in SCF to help update the cluster centroid of each pseudo cluster in the memory bank.
Using this approach, we not only tighten the clusters but also encourage clusters of the same individual in different clothing to be influenced by the current clustered groups and pulled closer toward them.
Second, we designed the SSF to deal with samples taken from front/back views. 
In SSF, we utilize a view classifier to identify sequences captured from front/back views. We then employ curriculum learning to gradually incorporate these sequences with those captured from other views.
Namely, the sequences are absorbed at a dynamic rate, relaxing the aggregate requirement for each cluster.
This approach enables us to re-assign pseudo labels for sequences captured from front/back views, thereby encouraging them to cluster with sequences captured from other views.
With our method, we gain a large recognition accuracy improvement compared to the baseline (with GaitSet~\cite{chao2019gaitset} backbone: NM + 3.1\%, BG + 8.6\%, CL + 9.7\%; with GaitGL~\cite{lin2021gait} backbone: BG + 1.1\%, GL + 17.2\% on CASIA-BN dataset~\cite{yu2006framework}\footnote{NM: normal walking condition, BG: carrying bags when walking, CL: walking with different coats.}).


To sum up, our contributions mainly lie in three folds:
\begin{itemize}
    \item[$\bullet$] We focus on Unsupervised Gait Recognition (UGR) using a cluster-based method with contrastive learning. Despite its practicality, it requires careful consideration. To address this task, we establish a baseline using cluster-level contrastive learning.
    
    \item[$\bullet$]  We deeply explore the characteristics of UGR, finding the two main challenges: clustering sequences with different clothes and with front/back views. To address these challenges, we propose a \textbf{S}elective \textbf{F}usion(SF) method. This method involves selecting potentially matched cluster/sample pairs to help them fuse gradually.
    
    \item[$\bullet$] Extensive experiments on three popular gait recognition benchmarks have shown that our method can bring consistent improvement over baseline, especially in walking in different coat conditions. 
\end{itemize}

\section{Related Work}\label{sec:relatedwork}

Gait recognition plays an important role in enhancing safety and security in the development of intelligent cities~\cite{zhang2024research}. Most existing gait recognition works are trained in a supervised manner, in which cross-cloth and cross-view labeled sequence pairs have been provided.
They mainly focus on learning more discriminative features~\cite{liao2020model,li2020jointsgait,chao2019gaitset,fan2020gaitpart,lin2021gait,9916067,9229117,9913216,10042966} or developing gait recognition applications in natural scenes~\cite{hou2022gait,das2023gait,9870842,9928336}.
However, obtaining labeled training pairs is challenging in real-world applications. Despite extensive research in gait recognition, further exploration of its practical applications is still needed.
In this work, we consider a practical setting and take one of the first steps toward achieving gait recognition without the need for labeled training datasets.

\subsection{Gait Recognition}

\noindent\textbf{Model-based method:} This kind of method encodes poses or skeletons into discriminative features to classify identities.
For example, PoseGait~\cite{liao2020model} extracts handcrafted features from 3D poses based on human prior knowledge. 
JointsGait~\cite{li2020jointsgait} extracts spatiotemporal features from 2D joints by GCN~\cite{yan2018spatial}, then maps them into discriminative space according to the human body structure and walking pattern.
GaitGraph~\cite{gaitgraph} uses human pose estimation to extract robust pose from RGB images, then encode the key points as nodes, and encode skeletons as joints in the Graph Convolutional Network to extract gait information.

\noindent\textbf{Appearance-based method:} This series of methods mostly input silhouettes, extracting identity information from the shape and walking postures.
GaitSet~\cite{chao2019gaitset} first extracts frame-level and set-level features from an unordered silhouette set, promoting the set-based method's development.
GaitPart~\cite{fan2020gaitpart} further includes part-level pieces of information, mining details from silhouettes.
In contrast, the video-based method GaitGL~\cite{lin2021gait} employs 3D CNN for feature extraction based on temporal knowledge.
Our method can be used in appearance-based unsupervised gait recognition.
%
%
In our framework, we adopt both the set-based method and the video-based method as the backbones to illustrate the generalization of our framework.

\noindent\textbf{Gait Recognition with Contrastive Learning:} The core idea of contrastive learning methods~\cite{chen2020simple, chen2021exploring, he2020momentum} is to construct effective positive and negative sample pairs through data augmentation and to design appropriate loss functions to optimize the model for learning useful data representations. Inspired by such methods as MoCo~\cite{he2020momentum} and SimCLR~\cite{chen2020simple}, GaitSSB~\cite{fan2023learning} proposes a self-supervised framework to learn general gait representations from large-scale unlabeled walking videos. GaitSSB treats each gait sequence as a single instance and aims to learn discriminative instance-level sequence features through contrastive learning. However, the current data augmentation methods are limited. More importantly, the positive pairs are often drawn from the same sequence, resulting in very similar positive sample pairs. This makes it difficult to simulate the variations caused by changes in clothing and camera viewpoints, thereby limiting the ability to provide effective supervisory signals that can guide the model to learn robust features.
To address these challenges, our method adopts a clustering strategy to group unlabeled data and generate pseudo-labels, allowing the model to learn representations based on cluster assignments. This clustering-based approach generates high-quality pseudo-labels and, unlike traditional contrastive methods where positive pairs are drawn from the same sequence, offers a more diverse and reasonable definition of positive and negative samples across different sequences. Defining these samples across multiple sequences proves more effective in addressing the challenges of cross-clothing and cross-camera scenarios in gait recognition tasks.

\subsection{Unsupervised Person Re-identification}
\textbf{Short-term Unsupervised Re-ID:}
Most fully unsupervised learning (FUL) Re-ID methods estimate pseudo labels for sequences, which can be roughly categorized into clustering-based and non-clustering-based methods. 
Clustering-based methods~\cite{zeng2020hierarchical,wang2021camera,chen2021ice,xuan2021intra,zhang2023camera,dai2021cluster} first estimate a pseudo label for each sequence and train the network with sequence similarity.
In contrast, non-clustering-based methods~\cite{lin2020unsupervised,wang2020unsupervised} regard each image as a class and use a non-parametric classifier to push each similar image closer and pull all other images further.
In total, the accuracy of most non-cluster-based methods does not exceed the latest cluster-based methods, so we use the latter to solve UGR.

At present, there are some typical algorithms in clustering-based methods.
BUC~\cite{lin2019bottom} utilizes a bottom-up clustering method, gradually clustering samples into a fixed number of clusters. Though there is a need for more flexibility, it is a good starting point. 
HCT~\cite{zeng2020hierarchical} adopts triplet loss to BUC to help learn hard samples. 
SpCL~\cite{ge2020self} introduces a self-paced learning strategy and memory bank, gradually making generated sample features closer to reliable cluster centroids. 
To alleviate the high intra-class variance inside a cluster caused by camera styles, CAP~\cite{wang2021camera} proposes cross-camera proxy contrastive loss to pull instances near their own camera centroids in a cluster. 
ICE~\cite{chen2021ice} further explores inter-instance relationships instead of using camera labels to compact the clusters with hard contrastive loss and soft instance consistency loss. 
IICS~\cite{xuan2021intra} also considers the difference caused by cameras, decomposing the training pipeline into two phases. First, it categorizes features within each camera and generates labels. Second, according to sample similarity across cameras, inter-camera pseudo labels will be generated based on all instances. These two stages train CNN alternately to optimize features.
Cluster-contrast~\cite{dai2021cluster} improves SpCL by establishing a cluster-level memory dictionary, optimizing and updating both CNN and memory bank at the cluster level.

On the contrary, the non-clustering-based methods mainly realize fully unsupervised Re-ID with similarity-based methods. 
SSL~\cite{lin2020unsupervised} predicts a soft label for each sample and trains the classification model with softened label distribution. 
MMCL~\cite{wang2020unsupervised} formulates FUL Re-ID as a multi-label classification task and classifies each sample into multiple classes by considering their self-similarity and neighbor similarity.

Inspired by the simple but elegant structure of Cluster-contrast~\cite{dai2021cluster}, we start from the Cluster-contrast framework to solve the UGR task.
Unsupervised~\cite{zeng2020hierarchical,dai2021cluster} and semi-supervised learning~\cite{chen2023class,chen2021multimodal} rely on modeling visual features (e.g., color, texture, shape) extracted from static images. Moreover, ReID methods typically depend on visual consistency across camera views, assuming that the same identity retains relatively stable appearance characteristics. However, gait recognition uses silhouette sequences as input, and it needs to identify features not only cross-view but also cross-clothes.

In summary, due to the differences in data modalities and the unique challenges faced by gait recognition, directly applying the cluster contrast method from~\cite{dai2021cluster} to the UGR task is not suitable.
As a solution, we design the SCF module with a support set that contains several cluster candidates the feature belongs to. Rather than pushing each feature toward only one cluster, we reduce the negative impact of erroneous gradients by minimizing incorrect associations. To tackle the challenge of clustering features from front/back views with those from other views, we also develop a novel method SSF to specifically help clustering these features from sparse views. 

\noindent\textbf{Long-term Unsupervised Re-ID:}
Gait recognition is a long-term task with cloth-changing, so long-term FUL Re-ID is more similar to UGR.
CPC~\cite{li2022unsupervised} uses curriculum learning~\cite{bengio2009curriculum} strategy to incorporate easy and hard samples and gradually relax the clustering criterion.
We do not use the same method in SCF, since each cluster mainly contains sequences with one cloth type, and it is better to pull clusters as a whole.
In contrast, we use curriculum learning in SSF to distinguishly deal with sequences in front/back views and gradually re-assign pseudo labels for those sequences.

\section{Our Method}
In this work, we propose a new task called Unsupervised Gait Recognition (UGR), which is practical when dealing with realistic unlabeled gait datasets.
In this section, we first formally define our technique.
Next, we will show our baseline based on Cluster-contrast~\cite{dai2021cluster}, trying to solve UGR with an unlabeled training set.
Then, we deeply research the problems faced by UGR and find two challenges to improve the accuracy: sequences in different clothes of the same person tend to form different clusters, and the sequences captured from front/back views are difficult to gather with other views.
Based on the two problems, we propose Selective Fusion to gradually merge cross-cloth clusters and sequences taken from front/back views to make samples of each.  

\begin{table}[ht]
\centering
\caption{The definition of important symbols}
\resizebox{\columnwidth}{!}{%
\begin{tabular}{c|c}
\toprule
Symbol & Definition \\ \midrule
$\mathcal{X}_u/\mathcal{X}_t$ & \begin{tabular}[c]{@{}c@{}}Unlabeled training dataset/\\ Labeled testing dataset\end{tabular} \\ \hline
$\mathcal{Y}_u/\mathcal{Y}'_u$ & The true/ pseudo label for training dataset \\ \hline
$\mathcal{Y}_t$ & The true label for testing dataset \\ \hline
$f_\theta$ & Gait Recognition backbone \\ \hline
$N$ & The total sequence number of the training dataset \\ \hline
$f_i$ & \begin{tabular}[c]{@{}c@{}}The feature extracted from the $i$-th \\ sequence of the training dataset  \end{tabular}  \\ \hline

$n$ & \begin{tabular}[c]{@{}c@{}} The number of neighbors KNN \\ searched for each sequence  \end{tabular} \\ \hline
$s_{up}$ & The similarity threshold KNN used for clustering \\ \hline
$Q$ & The number of clusters \\ \hline
$\mathcal{C}_k$ & The $k$-th cluster centroid \\ \hline
$\mathcal{M}$ & The memory bank  \\ \hline
$b$ & The mini-batch extracted from pseudo clusters each iteration  \\ \hline
$q$ & One query contained in $b$ \\ \hline
$\tau$ & The temperature hyper-parameter \\ \hline
$m$ & The momentum hyper-parameter \\ \hline
$\mathcal{L}_{q_{CNCE}}$ & The ClusterNCE Loss of query $q$ \\ \hline
$\mathcal{C}_{+}$ & The positive cluster centroid \\ \hline

$f_{i_{CA}}$ & The feature of $i$-th cloth augmented sequence \\ \hline
$\mathcal{C}_{ak}$ & The $k$-th adjusted cluster centroid \\ \hline

$\mathcal{S}_k$ & The support set of the $k$-th cluster \\ \hline
$a$ & The number of pseudo ids $\mathcal{S}_k$ contains.  \\ \hline
$c_{low}$ & The lower bound to judge FVC \\ \hline
$s_{c}/s_{o}$ & \begin{tabular}[c]{@{}c@{}}The current/initial similarity bound \\ when re-assign pseudo labels for sequences in FVC  \end{tabular}  \\ \hline
$\lambda/\lambda_{base}$ & \begin{tabular}[c]{@{}c@{}} Each epoch ratio/base ratio to merge \\ extreme view sequences in each epoch\end{tabular} \\ \hline
$\mathcal{C}_n/\mathcal{C}_o$ & The number of new or old clusters in each epoch \\ \hline

\end{tabular}%
}

\label{tab:symbol}
\end{table}

\subsection{Problem Formulation}

To formulate the unsupervised learning, we first define an unlabeled training dataset, denoted as $\mathcal{X}_u=\{x_{1}, x_2, ..., x_N\}$, including diverse conditions such as changes in clothing and viewpoint, where $N$ is the total sequence number.
%
%
We want to train a gait recognition backbone $f_\theta$ to classify these sequences according to their similarity. By clustering the features, we generate pseudo-labels $\mathcal{Y}_u$ for the training dataset. Following this, specialized modules are employed to gradually merge samples from different clothes and views.
During evaluation, $f_\theta$ will extract features from a labeled test dataset $\{\mathcal{X}_t, \mathcal{Y}_t\}$ and the gallery will rank according to their similarity with the probe, then we gain the rank-1 accuracy for each condition and each view.
We aim to train $f_\theta$ and gain the best performance on $\mathcal{X}_t$.


\subsection{Proposed Baseline}
We modified Cluster-contrast~\cite{dai2021cluster} to build our baseline framework. Since a pre-trained model is required to initialize $f_\theta$, we first pre-train the backbones on a large gait recognition dataset OU-MVLP~\cite{takemura2018multi} and then load it when training on the unlabeled dataset. The training pipeline for the unlabeled dataset can be summarized as follows: 
%
%
%

1) At the beginning of each epoch, we first use $f_\theta$ to extract features from each sequence in the training dataset in parts, which has been sliced by Horizontal Pyramid Matching (HPM)~\cite{chao2019gaitset} horizontally and equally. 
Then we concatenate all the parts to an embedding and regard it as a sequence feature to participate in the following process, denoted as $f_i, i \in \{1,2,..., N\} $. 
In this way, we can consider each sequence's features in parts and details.

2) We adopt KNN~\cite{fix1989discriminatory} to search $n$ neighbors for each sequence in feature space and calculate the similarity distances between each other. 
Then InfoMap~\cite{rosvall2008maps} is used to cluster $f_i$ with a similarity threshold $s_{up}$, and predict a pseudo label $\mathcal{Y}'_u= \{y'_{1},y'_{2},...,y'_{N} \}$ for each sequence.
When mapping features with a pre-trained model, each sequence tends to be mapped closer and not well separated.
So we tighten $s_{up}$, aiming to separate each subject into a single cluster.

3) With the pseudo labels, we compute the centroid of each cluster $\mathcal{C}_k, k \in \{0,1,...,Q\}$, $Q$ is the number of clusters, and then initialize a memory bank $\mathcal{M}$ at the cluster-level to store these centers $\mathcal{M} = \{\mathcal{C}_0, \mathcal{C}_1, ..., \mathcal{C}_Q\}$.

4) During each iteration, a mini-batch $b$ will be randomly selected from pseudo clusters, and during gradient propagation, we update the backbone with a ClusterNCE loss~\cite{dai2021cluster}.
\begin{equation}
\label{ClusterNCE}
    \mathcal{L}_{q_{CNCE}}=-\log \frac{\exp \left( q\cdot \mathcal{C}_+/\tau \right)}{\sum_{k=1}^Q{\exp}\left( q\cdot \mathcal{C}_k/\tau \right)}
\end{equation}
for each query feature $q$ extracted from the samples in the mini-batch, we calculate its similarity with the positive cluster centroid $\mathcal{C}_+$ it belongs to, $\mathcal{C}_k$ is the $k$-th cluster centroid, $\tau$ is the temperature hyper-parameter. Here we use all the query features in the mini-batch to calculate $\mathcal{L}_{q_{CNCE}}$.

5) Also, we update the centroids in the memory bank $\mathcal{M}$ that the queries belong to at a cluster level.
\begin{equation}
   \forall q \in \mathcal{C}_k, \mathcal{C}_k\gets m\mathcal{C}_k+(1-m)q
   \label{eq:momentum}
\end{equation}
$m$ is the momentum hyper-parameter used to update the centroids impacted by the batch. Each feature is responsible for updating the cluster centroids it belongs to. Additionally, $m$ serves as a crucial factor in determining the pace of cluster memory updates, whereby a higher value results in a slower update. This momentum value directly influences the consistency between the cluster features and the most recently updated query instance feature.

With this pipeline, we can initially realize UGR.
However, some defects still prevent further improvement in both cross-cloth and cross-view situations.
First, each cloth condition of a subject has been separated from each other, making it hard to group them together into a single class.
This is because of the large intra-class diversity within each subject when the identity changes cloth and subtle inter-class variance between different persons when clothes types of different subjects are similar.
For example, NM and CL of one person are less similar in appearance to NM of other persons, leading to the intra-similarity being smaller than the inter-similarity.
Second, some sequences in front/back views (such as $0^{\circ}/018^{\circ}/162^{\circ}/180^{\circ}$) cannot correctly gather with sequences in other views, but tend to be confused with front/back views sequences of other subjects.
This is because sequences in these views lack enough walking patterns, so the model can only use the shape information to classify these sequences. With more similarity in appearance, sequences of different subjects in these views tend to be classified together.
Necessary solutions need to be considered to help the framework solve the two problems.

\subsection{Proposed Method}

\begin{figure*}[t]
	\centering	 
	\includegraphics[width=\linewidth]{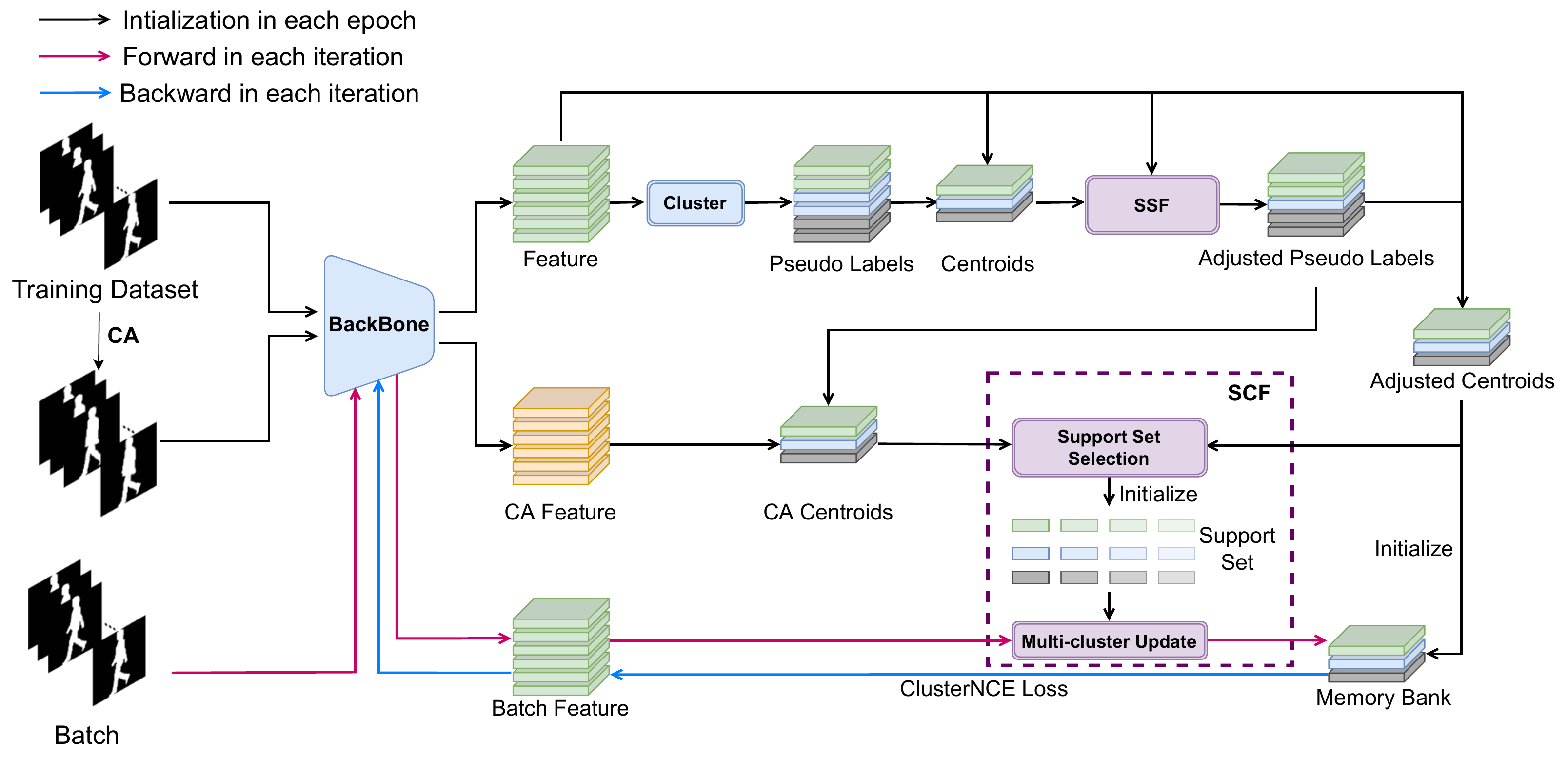}
	\caption{Overview of the framework with Selective Fusion. The upper two branches generate pseudo labels and initialize a memory bank at the start of each epoch. The lower branch accepts mini-batch extracted from pseudo clusters and calculates ClusterNCE Loss with Memory Bank to update it and the backbone in each iteration. CA is the \textit{Cloth Augmentation} method. InfoMap is employed in the Cluster module. SCF means \textit{Selective Cluster Fusion}. SSF represents \textit{Selective Sample Fusion}. In the Support set, the darker the color, the higher the similarity with the target cluster (Best viewed in color).}
	\label{fig:structure}
\end{figure*}

To tackle the problems we pointed out for UGR, we developed Selective Fusion, containing Selective Cluster Fusion (SCF) and Selective Sample Fusion (SSF) to solve the two drawbacks separately.
The framework of our method is in Figure~\ref{fig:structure}, and the pseudo-code is shown in Algorithm~\ref{our}.

\begin{algorithm}[t]
\caption{The training procedure of Selective Fusion}
\label{alg:algorithm}
\textbf{Input}:  $\mathcal{X}_u$;  $f_\theta$
\begin{algorithmic}[1]  
\REQUIRE Epoch number; iteration number; batch size; \\
hyper-parameters: $M/s_{up}/\tau/m/\mathcal{C}_{low}/\lambda_{base}/k/\mathcal{S}_{i}$
\FOR {epoch in range(0, epoch number+1) }
    \STATE {Apply CA to $\mathcal{X}_u$, get the augmented $\mathcal{X}_u$;}
    \STATE {Extract $f_n$ from $\mathcal{X}_u$ by $f_\theta$;}
    \STATE {Extract $f_{CA}$ from augmented $\mathcal{X}_u$ by $f_\theta$;}
    \STATE {Generate $\mathcal{Y}'_u$ for $f_n$}
    \STATE {Calculate the pseudo centroids with $\mathcal{Y}'_u$ and $f_n$;}
    \STATE {Generate adjusted $\mathcal{Y}'_u$ in SSF;}
    \STATE {Generate adjusted centroids with adjusted $\mathcal{Y}'_u$ and $f_n$;}
    \STATE {Initialize Memory Bank with adjusted centroids;}   
    \STATE {Generate support set in Support Set Selection Module with the centroids of $f_{CA}$ and $f_n$;}
    \FOR {iter in range(0, iteration number+1) }
    \STATE {Extract mini-batch from pseudo clusters;}
    \STATE {Extract sequences feature from batch;}
    \STATE {Calculate ClusterNCE loss with Memory Bank according to Eq.\textcolor{red}{~\ref{ClusterNCE}};}
    \STATE {Update the backbone;}
    \STATE {Update the Memory Bank according to Eq.\textcolor{red}{~\ref{momentum}};}
    \ENDFOR
\ENDFOR \\
\textbf{Output}: $f_\theta$
\end{algorithmic}
\label{our}
\end{algorithm}

In our method, at the beginning of each epoch, we use a Cloth Augmentation (CA) method to randomly generate an augmented variant for each sequence in the training dataset, then put them into the same backbone to extract features, named $f_{i}$ and $f_{i_{CA}}$.
Second, after getting the original pseudo labels generated by InfoMap, we use SSF to adjust the pseudo labels and then apply them to $f_{i}$ and $f_{i_{CA}}$ to get their adjusted centroids.
The adjusted centroids $\mathcal{C}_{ak}$ are used to initialize the memory bank $\mathcal{M}$.
Third, we use a support set selection module to generate a support set for each cluster, which will be used in the multi-cluster update strategy during back-propagation to help update the memory bank.
The support set selection module and multi-cluster update strategy are the two components of our SCF.

Next, we will introduce how we implement our Cloth Augmentation, SCF, and SSF methods.

\subsubsection{Cloth Augmentation}

\begin{figure}[ht]
	\centering	 
	\includegraphics[width=\linewidth]{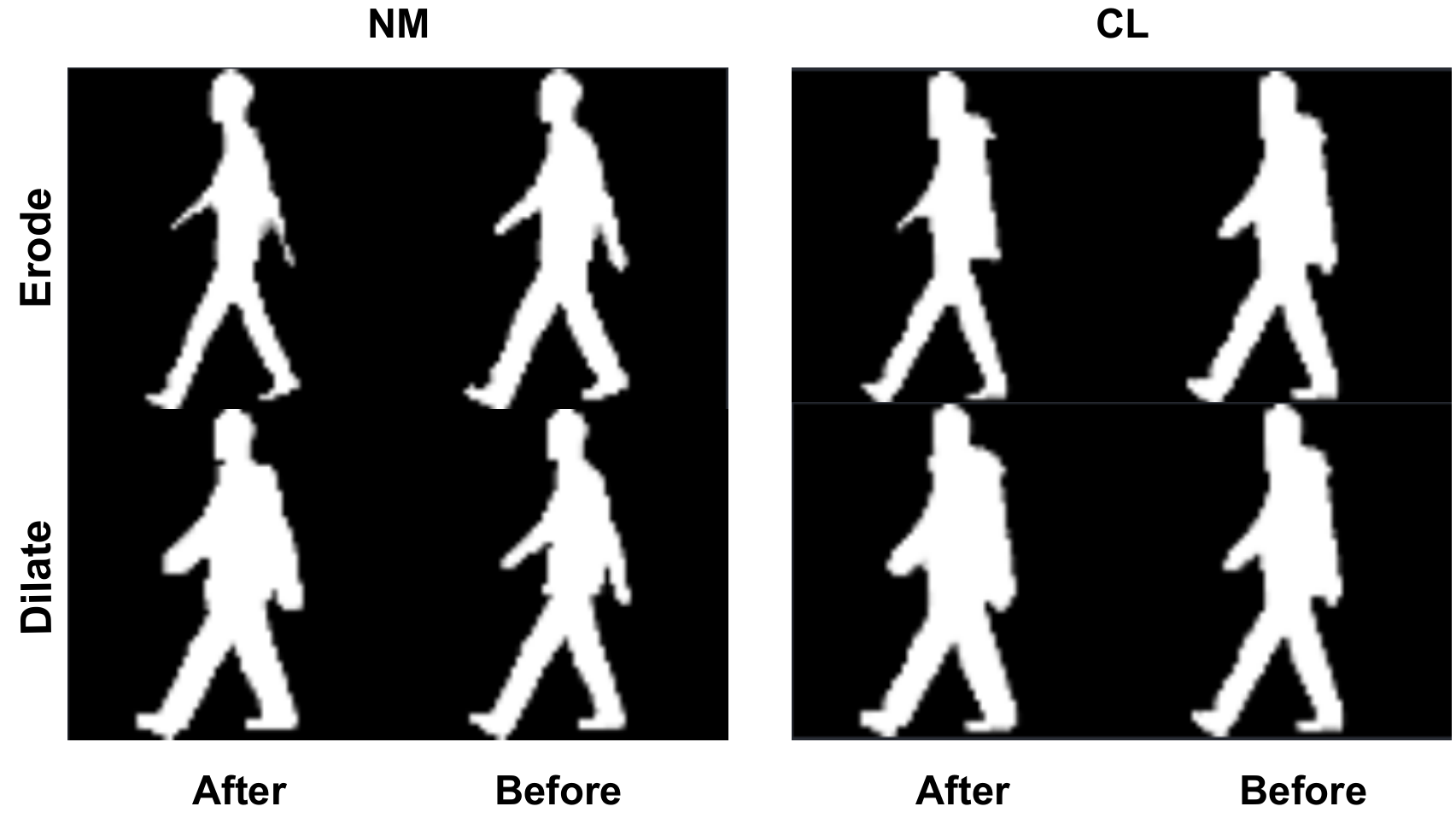}	 	
	\caption{The visualization of data augmentation on NM and CL conditions. Cloth Augmentation can simulate the potential appearance in different conditions of the same person.}
	\label{fig:DA}
\end{figure}

The cloth augmentation is conducted for each sequence in the training set to explicitly get a fuse direction for each cluster, which simulates the potential clusters in other conditions belonging to the same person.
Currently, the cloth augmentation methods we use are targeted for silhouette datasets, which the majority of algorithms work on.
We randomly dilate or erode the upper/bottom/whole body in the whole sequence with a probability of 0.5\footnote{The kernel size for upper part: $5 \times 5$, lower part: $2 \times 2$}, forming six cloth augmentation types.
Also, the upper/middle/bottom has a dynamic edited boundary\footnote{The boundary selected from upper bound: [14, 18], middle bound: [38, 42], bottom bound:[60, 64] for $64 \times 64$ silhouettes}, adding more variance to the augmentation results.
Here we visualize some cloth augmented results in Figure~\ref{fig:DA}. When dilating NM, the sequence can simulate its corresponding CL condition, and when eroding CL, the appearance of subjects can be regarded as in NM condition.
It is indeed that in real-world cloth-changing situations, the clothes have more diversity, only dilating or eroding cannot fully simulate all the situations.
Currently, we first consider the simple cloth-changing situation that walking with or without coats, which is also the cloth-changing method in CASIA-B~\cite{yu2006framework} and Outdoor-Gait~\cite{song2019gaitnet} dataset, to prove our method is valid.
In real application, automatic cloth augmentation methods can be employed, to automatically search other cloth augmentation methods, such as sheer the bottom part of the silhouettes to simulate wearing a dress and adding an oval above the head to simulate wearing a hat, which is another promising research direction in the future research.
In this work, we use clothing augmentation via dilation or erosion in our method to provide the opportunity to let the cross-cloth sequences have the chance to be closer to each other. 
This will facilitate our method of utilizing the close chance to further pull the clusters near through the support set and multi-cluster update strategy.

%
\subsubsection{Selective Cluster Fusion}
SCF aims to pull clusters in different clothes belonging to the same subject closer.
It comprises two parts, a support set selection, which is used to generate a support set $\mathcal{S}_k$ for each cluster, aiming to find potential candidate clusters in different clothes, and a multi-cluster update strategy, aiming to decrease the distance between candidate clusters in the support set.

\textbf{Support Set:} The input of the support set selection module is the centroids of $\mathcal{C}_{ak}$ and $\mathcal{C}_{k}$.
By calculating the similarity between the $k$-th Cloth Augmented centroid $\mathcal{C}_{ak}$ with all the original centroids contained in $\mathcal{M}$, we can get a rank list with these pseudo labels, ranging from highest to lowest according to their similarity distances.
We select the top $a$ ids in each rank list, and the first id we set is the cluster itself, formulating the support set $\mathcal{S}_k = {id_0,id_1,...,id_a}$.
With the support set, we can concretize the optimization direction when pulling NM and CL together because $f_{i_{CA}}$ can be seen as the cross-cloth sequences in reality to some extent.
With the explicit regulation, we will not blindly pull a cluster close to any near neighbor.

\textbf{Multi-cluster update strategy:} When updating the memory bank during backpropagating, we use the support set in the multi-cluster update strategy.
Knowing which clusters are the potential conditions of one person, the new strategy can be formulated as follows:
\begin{equation}
\label{momentum}
   \forall \mathcal{C}_{ak} \in \mathcal{S}_k, \forall q \in \mathcal{C}_{ak}, \mathcal{C}_{ak} \leftarrow m \mathcal{C}_{ak} + (1-m) q
\end{equation}
All the candidate clusters in $\mathcal{S}_k$ need to participate in updating the memory bank.
By forcing the potential conditions to fuse, we can make the mini-batch influence clusters and clusters in the support set, compressing the distance between cross-cloth pairs.


\subsubsection{Selective Sample Fusion}

Seeing that walking postures are absent in front/back views, sequences taken from these views have less feature similarity with features extracted from other views.
So they cannot be appropriately gathered into their clusters like other views, and tend to mix up with sequences of other identities captured from front/back views.
If we pull all the clusters towards their candidate clusters in the support set, those clusters mainly composed of sequences taken from front/back views will further gather with clusters with the same view condition, making the situation worse. 
To deal with clusters in this condition, we design SSF, in which we use curriculum learning to gradually re-assign pseudo labels to sequences in front/back views, forcing them to fuse with samples taken from other views progressively before conducting the SCF method.

\textbf{View classifier :} Specifically, we first train a view classifier on OU-MVLP, classifying whether the sequence is in the front/back view.
The view classifier structure we set is as same as the GaitSet structure we used, and we add the BNNeck behind. 
We assign the label 1 for sequences in $0^{\circ}/180^{\circ}$ and assign the label 0 for sequences in other views to train the view classifier. 
We train the view classifier on OU-MVLP to gain view knowledge, and when we have prior knowledge of sequences' view, we can quickly identify which clusters generated by our framework are composed of sequences taken from front/back views.

Indeed, it is true that when adapting the view classifier to other datasets, there may be appearance domain gaps and view classification gaps between different datasets, as other datasets may not label views in the same way as OUVMLP, and some datasets may not provide view labels at all.
However from our observations, we found the view classification task is not a hard task, the view classifier can already identify most sequences with fewer walking postures.
To further solve the problem that some sequences haven't appropriately assigned the right view label, we set a threshold to relax the requirement when clustering.
The criterion is that only when the number of sequences in the front/back view in one cluster is larger than the threshold $c_{low}$, we consider the cluster as Front/Back View Clusters (FVC).
%
By dissolving FVC, we calculate the similarity between each sequence in it with other centroids of non-FVC, and if the similarity is larger than $s_{c}$, we re-assign the nearest non-FVC pseudo label for the sequence.
Therefore, even if not all sequences with front or back views are correctly identified, aligning most of these sequences closer to other views can help reduce the feature disparity between these perspectives and other views.
And with the training process of curriculum learning, the clusters are tighter. The sequences identified with front/back views will also push the sequences that were misidentified closer to their cluster center automatically since they have appearance similarity. 
More intelligent methods can be developed in future research to reduce the dependence of the model on the prior information.

\textbf{Curriculum Learning:} However, we do not incorporate all the sequences in FVC at the same time, instead, we utilize Curriculum Learning~\cite{bengio2009curriculum} to fuse them progressively by enlarging the $s_{c}$ during each epoch:
\begin{equation}
    s_{c} = s_{o} - \lambda \times epoch\_number
\end{equation}
At the first epoch, $s_{o}$ is high, and only the similarity between sequences in FVC and centroids of other non-FVC higher than $s_{i}$, can be assigned new pseudo labels. 
Otherwise, they will be seen as outliers and cannot participate in training.
During training, the criterion gradually increases with a speed of $\lambda$, which allows for the gradual incorporation of new knowledge.
%
%
In response to this, we propose to set $\lambda$ adaptively~\cite{hou2019learning}:
\begin{equation}
\lambda=\lambda_{base} \left| \frac{\mathcal{C} _n}{\mathcal{C} _o} \right|
\end{equation}
where $\left|\mathcal{C}_n\right|$ and $\left|\mathcal{C}_o\right|$ is the number of new and old clusters in each epoch, $\lambda_{base}$ is a fixed constant for each dataset. 
Since more clusters are fused in each phase, $\lambda$ increases as the ratio of the number of new clusters to that of old clusters increases.
%

\subsection{Training Strategy}
\begin{figure}[htbp]
	\centering	 
	\includegraphics[width=\linewidth]{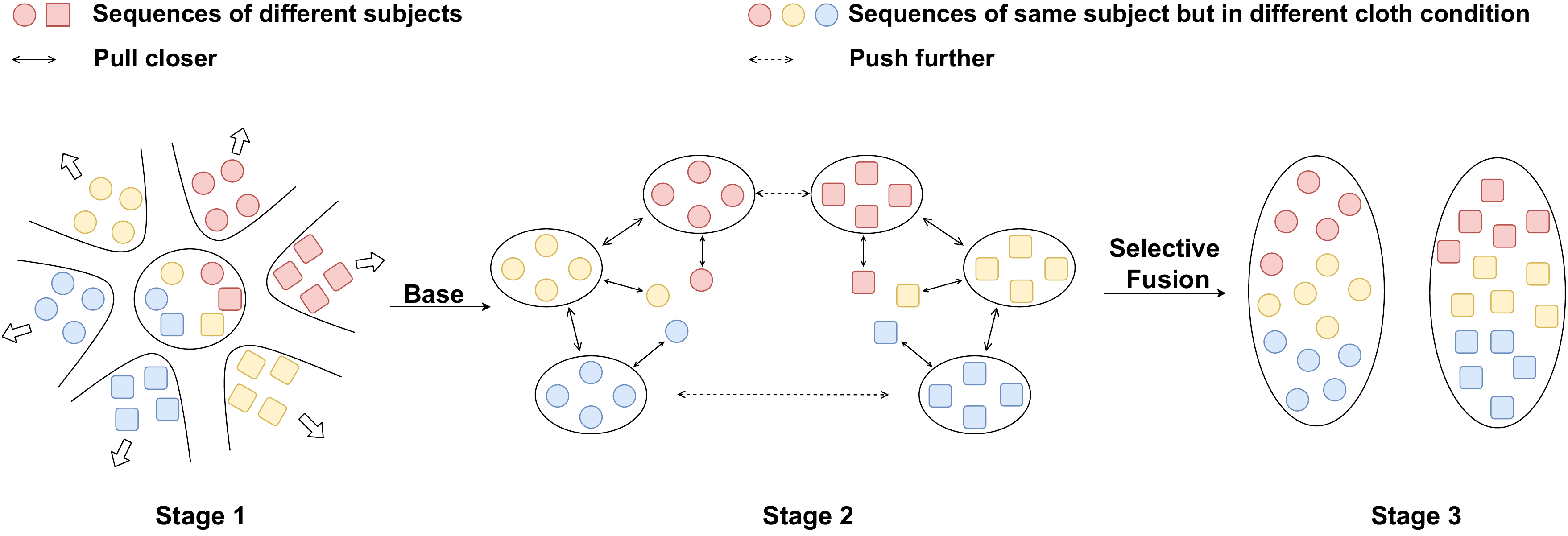}	 	
	\caption{The three stages in our training strategy. First, with narrowed features extracted by the pre-trained model, we first adopt our baseline to separate them further. Then Selective Fusion is used to fuse matched clusters and samples together. Finally, we gain clusters with different clothes and views. The base is the \textit{Baseline}. Each type in a different color indicates \textit{each subject in a different cloth condition.} }
	\label{fig:stage}
\end{figure}

Overall, Selective Fusion can make separated conditions and scattered sequences taken from front/back views fuse tighter.
Here we show our training strategy and represent the feature distribution of each training phase in Figure~\ref{fig:stage}.
Our training strategy encompasses three stages.
At first, the features extracted by the pre-trained model have a tendency to cluster together, making it hard to distinguish between them.
We adopt our baseline to separate these sequences with a strict criterion, making each cluster gathered according to their similarity.
Second, we apply Selective Fusion to fuse different conditions of the same person and gradually merge sequences in front/back views with sequences in other views.
Finally, we get the clusters with all the cloth conditions and views.
%

\begin{table*}[h]
\centering
\caption{The parameters used in unsupervised learning on OU-MVLP, CASIA-BN, Outdoor-Gait dataset, and GREW.}
\setlength{\tabcolsep}{1pt}
\begin{tabular}{ccccccc}
\hline
Param                           & Backbone  &OU-MVLP & CASIA-BN                 & Outdoor-Gait   &GREW  \\  
\hline
\multirow{2}{*}{Model Channel}  & GaitSet   & -        & (32, 64, 128)(128, 256)  & (32, 64, 128)(128, 256) &(32, 64, 128,256)(256, 256)\\
                                & GaitGL    & -        & (32, 64, 128)(128, 128)  & -                       &- \\
Batch Size                      & Both      & (32, 16) & (8, 16)                  & (8, 8)                  & (8,16)                \\
Weight Decay                    & Both      & 5e-4     & 5e-4                     & 5e-4                    &5e-4                  \\
Start LR                        & Both      & 1e-1     & 1e-4                     & 1e-4                    &1e-4                \\
Milestones                      & Both      & -        & {[}3.5k, 8.5k{]}         & {[}3.5k, 8.5k{]}        & {[}3.5k, 8.5k{]}      \\
Epoch                           & Both      & -        & Baseline: 50, SF: 50     & Baseline: 50, SF: 50    & Baseline: 50, SF: 50  \\
Iteration                       & Both      & -        & Baseline: 50, SF: 100    & Baseline: 50, SF: 100   &Baseline: 1000, SF: 1000 \\
\multirow{2}{*}{\begin{tabular}[c]{@{}l@{}}Upper bound Milestones\end{tabular}} 
                                 & GaitSet  &{[}50k, 100k, 125k{]}      & {[}20k, 40k, 60k, 80k{]} & {[}10k, 20k, 30k, 35k{]} & {[}10k, 20k, 30k, 35k{]}\\
                                & GaitGL    & {[}150k, 200k, 210k{]}    & {[}70k, 80k{]}           & -  &-\\ \hline
\end{tabular}%
\label{unsupervise_param}
\end{table*}
\section{Experiments}
Our methods can be employed in appearance-based methods. 
For simplicity, we take silhouette sequences as input since they are more robust when datasets are collected in the wild.
To demonstrate the effectiveness of our framework, we apply our methods to two existing backbones: GaitSet~\cite{chao2019gaitset} and GaitGL~\cite{lin2021gait} to help them train with unlabeled datasets. 
We also compare our method with upper bound which is trained with the ground truth label, and with the baseline, which is also trained without supervision.
All methods are implemented with PyTorch~\cite{paszke2019pytorch} and trained on TITAN-XP GPUs.

\subsection{Datasets} 
We train and test the performance of our method on three popular datasets, CASIA-BN~\cite{yu2006framework}, Outdoor-Gait~\cite{song2019gaitnet} and GREW~\cite{lin2014effects}.
\subsubsection{CASIA-BN} 
The original CASIA-B is a useful dataset with both cross-view and cross-cloth sequence pairs. 
It consists of 124 subjects, having three walking conditions: \textbf{normal walking} (NM\#01-NM\#06), \textbf{carrying bags} (BG\#01-BG\#02), and \textbf{walking with different coats} (CL\#01-CL\#02).
Each walking condition contains 11 views distributed in $[0^{\circ},180^{\circ}]$. We employ the protocol in the previous research~\cite{chao2019gaitset,fan2020gaitpart}.
%
During the evaluation, NM\#01-NM\#04 are the gallery, NM\#05-NM\#06, BG\#01-BG\#02, CL\#01-CL\#02 are the probe.
Due to the coarse segmentation of CASIA-B, we collected some pedestrian images and trained a new segmentation model to re-segment CASIA-B, and gain CASIA-BN.

\subsubsection{Outdoor-Gait}
This dataset only has cross-cloth sequence pairs.
With 138 subjects, Outdoor-Gait contains three walking conditions: normal walking (NM\#01-NM\#04), carrying bags (BG\#01-BG\#04), and walking with different coats (CL\#01-CL\#04). 
There are three capture scenes (Scene\#01-Scene\#03), however, each person only has one view ($90^{\circ}$).
69 subjects are used for training and the last 69 subjects for tests.
During the test, we use NM\#01-NM\#04 in Scene\#03 as a gallery and all the sequences in Scene\#01-Scene\#02 as probes in different conditions.
\subsubsection{GREW}
To the best of our knowledge, 
GREW~\cite{lin2014effects} is the largest gait dataset in real-world conditions. The raw videos are gathered from 882 cameras in a vast public area, encompassing nearly 3,500 hours of 1,080×1,920 streams. Alongside identity information, some attributes such as gender, 14 age groups, 5 carrying conditions, and 6 dressing styles have been annotated as well, ensuring a rich and diverse representation of practical variations.
Furthermore, this dataset includes a train set with 20,000 identities and 102,887 sequences, a validation set with 345 identities and 1,784 sequences, and a test set with 6,000 identities and 24,000 sequences. In the test phase, we strictly follow the official test protocols.
\setlength{\tabcolsep}{5pt}
\begin{table*}[ht]
\centering
\caption{The rank-1 accuracy (\%) on CASIA-BN for different probe views excluding the identical-view cases. For evaluation, the sequences of NM\#01-NM\#04 for each subject are taken as the gallery. The probe sequences are divided into three subsets according to the walking conditions (\textit{i.e.} NM, BG, CL). SF is our \textit{Selective Fusion Method}. CC is the \textit{Cluster-contrast framework we followed}. \textcolor{BrickRed}{Red} indicates the upper bound with supervised learning. \textcolor{RoyalBlue}{\textbf{Blue}} indicates the improvements on sequences in front/back views. \textbf{Bold} indicates the total improvements on different conditions.}
\begin{tabular}{ccccccccccccccc}
\toprule
\multirow{2}{*}{Backbone} & \multirow{2}{*}{Condition} & \multirow{2}{*}{Method} & \multicolumn{11}{c}{Probe View} & \multirow{2}{*}{Average} \\ \cline{4-14}
 &  &  & $0^{\circ}$ & $18^{\circ}$ & $36^{\circ}$ & $54^{\circ}$ & $72^{\circ}$ & $90^{\circ}$ & $108^{\circ}$ & $126^{\circ}$ & $134^{\circ}$ & $162^{\circ}$ & $180^{\circ}$ &  \\ \midrule
\multirow{12}{*}{GaitSet} & \multirow{4}{*}{NM} & Upper & 90.5 & 98.1 & 99.0 & 96.9 & 93.5 & 91.0 & 94.9 & 97.8 & 98.9 & 97.2 & 83.4 & \textcolor{BrickRed}{94.7} \\
 &  & Pretrain & 59.6 & 72.9 & 80.1 & 77.4 & 67.4 & 58.8 & 63.9 & 72.4 & 79.2 & 62.2 & 42.7 & 67.0 \\
 &  & Base (CC) & 77.7 & 92.0 & 94.7 & 92.8 & 88.4 & 83.8 & 86.2 & 91.2  & 93.0 & 90.5 & 69.4 & 87.2  \\
 &  & Ours (SF) & \textcolor{RoyalBlue}{\textbf{85.2}} & 93.6 & 96.4 & 93.8 & 90.0 & 84.6 & 89.6 & 92.3 & 96.9 & 93.2 & \textcolor{RoyalBlue}{\textbf{77.4}} & \textbf{90.3} \\ \cline{2-15} 
 
 & \multirow{4}{*}{BG} & Upper & 86.1 & 94.1 & 95.9 & 90.7 & 84.2 & 79.9 & 83.7 & 87.1 & 94.0 & 93.8 & 78.0 & \textcolor{BrickRed}{88.0} \\
 &  & Pretrain & 48.0 & 51.9 & 58.1 & 54.2 & 52.2 & 45.8 & 47.6 & 49.9  & 53.5 & 45.5 & 36.4 & 49.2 \\
 &  & Base (CC) & 70.4 & 81.5 & 84.0 & 78.0 & 74.3  & 67.0 & 71.9 & 73.7  & 77.3 & 76.5 & 62.5 & 74.3 \\
 &  & Ours (SF) & \textcolor{RoyalBlue}{\textbf{78.5}} & \textcolor{RoyalBlue}{\textbf{88.3}} & 89.8 & 88.0 & 83.5 & 76.4 & 80.5 & 83.5 & 85.8 & \textcolor{RoyalBlue}{\textbf{84.8}} & \textcolor{RoyalBlue}{\textbf{72.6}} & \textbf{82.9} \\ \cline{2-15} 
 & \multirow{4}{*}{CL} & Upper & 65.2 & 79.3 & 84.4 & 81.0 & 77.9 & 74.1 & 75.7 & 79.2 & 81.5 & 73.2 & 47.5 & \textcolor{BrickRed}{74.5} \\
 &  & Pretrain & 9.8 & 10.7 & 14.4 & 17.3 & 16.1 & 13.6 & 15.2 & 14.8 & 13.1  & 7.8 & 6.8 & 12.7 \\
 &  & Base (CC) & 27.7 & 32.4 & 37.2 & 37.6 & 33.0 & 29.2 & 32.0 & 32.2  & 32.3 & 27.8 & 20.7 & 31.1 \\
 &  & Ours (SF) & \textcolor{RoyalBlue}{\textbf{33.1}} & \textcolor{RoyalBlue}{\textbf{41.6}} & 46.4 & 47.6 & 46.7 & 41.2 & 44.7 & 43.3 & 45.6 & \textcolor{RoyalBlue}{\textbf{35.8}} & 22.5 & \textbf{40.8} \\ \midrule

\multirow{12}{*}{GaitGL} & \multirow{4}{*}{NM} & Upper & 94.2 & 97.5 & 98.7 & 96.7 & 95.1 & 92.9 & 95.9 & 97.9 & 99.0 & 98.0 & 87.1 & \textcolor{BrickRed}{95.7} \\
 &  & Pretrain & 66.2 & 79.9 & 85.3 & 84.9 & 73.5 & 65.8 & 71.8 & 81.7 & 85.8 & 79.1 & 51.0 & 75.0 \\
 &  & Base (CC) & 83.7 & 94.7 & 96.4 & 93.2 & 88.2 & 85.1 & 87.6 & 91.7 & 95.5 & 94.3 & 70.8 & 89.2 \\
 &  & Ours (SF) & 84.2 & 95.7 & 96.2 & 94.7 & 89.9 & 87.0 & 89.0 & 91.8  & 96.7 & 93.2 & 63.6 & \textbf{89.3} \\ \cline{2-15} 
 & \multirow{4}{*}{BG} & Upper & 88.6 & 96.2 & 96.6 & 94.2 & 91.0 & 85.8 & 91.0 & 94.4 & 97.0 & 95.3 & 76.6 & \textcolor{BrickRed}{91.5} \\
 &  & Pretrain & 53.4 & 66.4 & 67.2 & 67.7 & 62.0 & 56.2 & 59.9 & 62.4 & 66.1 & 64.6 & 43.4 & 60.8 \\
 &  & Base (CC) & 74.6 & 88.7 & 87.6 & 85.0 & 83.2 & 79.3 & 80.1 & 83.0 & 87.4 & 86.9 & 63.0 & 81.7 \\
 &  & Ours (SF) & 75.8 & 90.1 & 91.6 & 87.5 & 83.9 & 80.2 & 83.0 & 85.0 & 89.7 & 87.9  & 56.0 & \textbf{82.8} \\ \cline{2-15} 
 & \multirow{4}{*}{CL} & Upper & 71.7 & 87.6 & 91.0 & 88.3 & 85.3 & 81.4 & 82.9 & 85.9 & 87.5 & 85.4 & 53.0 & \textcolor{BrickRed}{81.8} \\
 &  & Pretrain & 18.5 & 25.4 & 28.7 & 29.9 & 28.4 & 23.6 & 25.7 & 24.9 & 24.3 & 21.7 & 11.8 & 23.9 \\
 &  & Base (CC) & 33.7 & 49.6 & 55.0 & 55.1 & 58.2 & 53.4 & 57.3 & 52.6 & 52.2 & 41.7 & 24.1 & 48.4 \\
 &  & Ours (SF) & \textcolor{RoyalBlue}{\textbf{53.6}} & \textcolor{RoyalBlue}{\textbf{71.1}} & 76.9 & 75.5 & 72.9 & 69.1 & 69.8 & 69.0 & 71.7 & \textcolor{RoyalBlue}{\textbf{60.7}} & \textcolor{RoyalBlue}{\textbf{31.2}} & \textbf{65.6} \\ \bottomrule
\end{tabular}

\label{tab:dataset_casiabn}
\end{table*}

%
\subsection{Implementation Details} 
In this section, we provide comprehensive details about the implementation of our method. 
In the Re-ID task~\cite{dai2021cluster}, ResNet-50~\cite{he2016deep}, pre-trained on ImageNet~\cite{deng2009imagenet}, is commonly used as the backbone for feature extraction.
Similar to unsupervised Re-ID methods, a unified pre-trained model is crucial for initialization in UGR. However, there currently does not exist a unified pre-trained model that can be used for the gait community. Here we first pre-train backbones on a large gait recognition dataset OU-MVLP~\cite{takemura2018multi}, and then load it when training on the unlabeled dataset. OU-MVLP, with its large dataset volume, is an ideal dataset for pre-train models. It contains 10,307 subjects. The sequences of each subject distributed in 14 views between $[0^{\circ},90^{\circ}]$ and $[180^{\circ},270^{\circ}]$, but only have normal walking conditions. We used 5153 subjects to pre-train backbones.

It is worth noting that the OU-MVLP dataset is only used in the pre-training stage and is not involved in the unsupervised learning stage. There are no cloth-changing pairs in this dataset, preventing the transfer of cross-cloth information. And our method focuses on tackling the challenging cross-cloth task by clustering features and learning intrinsic information based on pseudo labels, without relying on any prior information from the pre-trained model. As a result, we abbreviate our method as Unsupervised Gait Recognition rather than a domain transformation.   Moreover, we believe that training a unified pre-trained model for the gait community is a promising direction for future work.

In Table~\ref{unsupervise_param}, we present the structure and optimization settings used for pre-training and unsupervised learning. Recognizing individuals on GREW proves more challenging due to its larger scale and diverse, real-world settings. Consequently, we employ convolutional layers with increased channel sizes (32, 64, 128, 256). For hyperparameters, we set $s_{up}=0.7$ for the baseline and $s_{up}=0.3$ for Selective Fusion to enlarge the boundary. The remaining parameters are set as $n=40$, $\tau=0.05$, $k=2$, $\lambda_{base}=0.005$, $c_{low} = 0.8$, and $s_{o} = 0.7$. Since the sequences in Outdoor-Gait are fewer, we change $n$ to 20 to avoid overfitting. The batch size is represented as $(B_S, B_T)$, where each mini-batch contains $B_S$ subjects. For each subject, $B_T$ sequences are sampled, and 30 frames are randomly selected from each sequence. Each frame is normalized to a size of $64\times44$. We use a cosine annealing strategy to update $m$ in Equation~\eqref{eq:momentum}, the strategy can be formulated as follows:
\begin{equation}
\label{momentum_strategy}
   m_t = m_{\text{min}} + \frac{1}{2} (m_{\text{max}} - m_{\text{min}}) \left(1 + \cos\left(\frac{t \pi}{T}\right)\right)
\end{equation}
where $m_t$ is the momentum at training step $t$, $m_{\text{min}}$ is the minimum momentum value, $m_{\text{max}}$ is the initial value, and $T$ represents the total number of training steps within a single epoch. As training progresses ($t$ increases), the momentum decreases following a cosine curve. We set $m_{\text{max}}=0.5$ and $m_{\text{min}}=0.1$.
\subsection{Benchmark Settings} To show the effectiveness, we define several benchmarks:

(1) \textit{Upper}. The upper bound reports the performance of each backbone trained with ground-truth labels.

(2) \textit{Pre-train}. The effect when directly applying the pre-trained model to the target dataset without fine-tuning.

(3) \textit{Base (CC)}. Fine-tuning the pre-trained model with our baseline framework implemented by Cluster-contrast.

(4) \textit{Ours (SF)}. The results of our proposed method.

\subsection{Performance Comparison}
%

\subsubsection{CASIA-BN}
The performance comparison on CASIA-BN is shown in Table~\ref{tab:dataset_casiabn}. 
We evaluate the probe in three walking conditions separately. 
Since our method aims to improve the rank-1 accuracy of CL and sequences in front/back views, \textbf{ we take the accuracy for CL as the main criteria}.
From the results, we can see that our method outperforms the baseline in the CL condition by a remarkable margin (GaitSet: CL + 9.7\%; GaitGL: CL + 17.2\%).
It indicates that our Selective Cluster Fusion method can properly identify the potential clusters of the same person with different cloth conditions and pull them together.
Moreover, sequences in $0^{\circ}/18^{\circ}/162^{\circ}/180^{\circ}$ also gained large improvement in both cloth conditions.
Selective Sample Fusion can gradually gather individual front/back samples that were excluded initially, by assigning them the same pseudo labels as the sequences in other views.
Although lacking walking postures, the sequences with front/back views can still provide useful information for identifying a particular person. 
It should be noted that the hyper-parameters used for GaitGL are as same as GaitSet and without specific adjustment, which shows the generalization of our method when applying to different backbones. 
Both cues indicate that our method is effective when dealing with cloth-changing and front/back views.

%

\subsubsection{Outdoor-Gait}
Although Outdoor-Gait does not consider cross-view data pairs, we can still verify the SCF method on this dataset and show the result with the GaitSet backbone in Table~\ref{tab:outdoor_gait}.
%

\setlength{\tabcolsep}{12pt}
\begin{table}[h]
\centering
\caption{The Rank-1 accuracy (\%) on Outdoor-Gait. When evaluation, we take NM\#1-NM\#4 in Scene\#3 as gallery and others as probe.}
\begin{tabular}{ccccc}
\hline
Backbone & Method & NM & BG & CL \\ \hline
\multirow{4}{*}{GaitSet} & Upper & \textcolor{BrickRed}{97.6} & \textcolor{BrickRed}{90.9} & \textcolor{BrickRed}{90.4} \\
 & Pretrain & 45.8 &  46.4 & 43.3 \\
 & Base (CC) & 84.8 & 66.5 & 62.9 \\
 & Ours (SF) & \textbf{90.0} & \textbf{73.7} & \textbf{71.9} \\ \hline
\end{tabular}
\label{tab:outdoor_gait}
\end{table}
\begin{table}[t]
\centering
\caption{The Rank-1 and Rank-5 accuracy (\%) on GREW. Trained on the GREW train set and evaluated on the test set.}
\setlength{\tabcolsep}{12pt}
\begin{tabular}{ccccc}
\hline
Backbone & Method &Rank-1 &Rank-5 \\ \hline
\multirow{4}{*}{GaitSet} & Upper & \textcolor{BrickRed}{48.4} & \textcolor{BrickRed}{63.6}  \\
 & Pretrain & 17.0 &  28.5  \\
 & Base (CC) & 18.3 & 30.4  \\
 & Ours (SF) & \textbf{20.2} & \textbf{32.0}\\ \hline
\end{tabular}
\label{tab:grew}
\end{table}
\begin{table*}[h]
\centering
\caption{Ablation study demonstrating the effectiveness of each module in our method on the CASIA-BN, Outdoor-Gait, and GREW datasets based on Rank-1 accuracy (\%)}
\setlength{\tabcolsep}{12pt}
\label{tab:each}
\begin{tabular}{cccccccc}
\hline
\multirow{2}{*}{Setting} & \multicolumn{3}{c}{CASIA-BN}                  & \multicolumn{3}{c}{Outdoor-Gait}              & \multirow{2}{*}{GREW} \\
                         & NM            & BG            & CL            & NM            & BG            & CL            &                       \\ \hline
Base                     & 87.2          & 74.3          & 31.1          & 84.8          & 66.5          & 62.9          & 18.3                  \\
Base + SSF               & 87.5          & 75.8          & 32.8          & 85.3          & 66.9          & 61.6          & 18.7                  \\
Base + SCF               & 83.3          & 75.9          & 39.9          & 89.1          & 73.6          & 71.9          & 19.7                  \\
Ours                     & \textbf{90.3} & \textbf{82.9} & \textbf{40.8} & \textbf{90.0} & \textbf{73.7} & \textbf{71.9} & \textbf{20.2}         \\ \hline
\end{tabular}%
\end{table*}
\begin{table*}[h]
\centering
\caption{The rank-1 accuracy (\%) on CASIA-BN for different probe views excluding the identical-view cases. The probe sequences are divided into three subsets according to the walking conditions (\textit{i.e.} NM, BG, CL).}
\setlength{\tabcolsep}{9pt}
\resizebox{\textwidth}{!}{%
\begin{tabular}{cccccccccccccc}
\toprule
\multirow{2}{*}{Backbone} & \multirow{2}{*}{Condition} & \multirow{2}{*}{Method} & \multicolumn{10}{c}{Probe View}  \\ \cline{4-14} 
 &  &  & $0^{\circ}$ & $18^{\circ}$ & $36^{\circ}$ & $54^{\circ}$ & $72^{\circ}$ & $90^{\circ}$ & $108^{\circ}$ & $126^{\circ}$ & $134^{\circ}$ & $162^{\circ}$ & $180^{\circ}$  \\ \midrule
\multirow{12}{*}{GaitSet} & \multirow{4}{*}{NM}& Base  & 77.7 & 92.0 & 94.7 & 92.8 & 88.4 & 83.8 & 86.2 & 91.2  & 93.0 & 90.5 & 69.4  \\ 
 &  & Base+SSF & \textcolor{RoyalBlue}{\textbf{80.1}} & \textcolor{RoyalBlue}{93.2} & 94.9 & 92.9 & 87.6 & 83.4 & 86.4 & 91.1 & 92.9 & \textcolor{RoyalBlue}{90.7} & \textcolor{RoyalBlue}{\textbf{69.9}}  \\ 
 &  & Base+SCF & 70.1 &89.3 &91.9 &91.8 &83.7 &82.1 &82.3 &90.8 &90.5 &81.6 &62.2   \\
&  & Ours (SF) & \textcolor{RoyalBlue}{\textbf{85.2}} & 93.6 & 96.4 & 93.8 & 90.0 & 84.6 & 89.6 & 92.3 & 96.9 & 93.2 & \textcolor{RoyalBlue}{\textbf{77.4}} \\ \cline{2-14} 
 
 & \multirow{4}{*}{BG} & Base & 70.4 & 81.5 & 84.0 & 78.0 & 74.3  & 67.0 & 71.9 & 73.7  & 77.3 & 76.5 & 62.5  \\
 &  & Base+SSF & \textcolor{RoyalBlue}{\textbf{72.9}} & \textcolor{RoyalBlue}{\textbf{84.0}} & 83.6 & 77.9 & 74.5 & 67.7 & 70.8 & 73.2 & 79.5 & \textcolor{RoyalBlue}{\textbf{78.1}} & \textcolor{RoyalBlue}{\textbf{71.3}} \\
 &  & Base+SCF &69.3 &79.5 &87.7 &79.8 &74.4 &74.5 &74.6 &77.7 &79.1 &76.1 &61.9 \\
 &  & Ours (SF) & \textcolor{RoyalBlue}{\textbf{78.5}} & \textcolor{RoyalBlue}{\textbf{88.3}} & 89.8 & 88.0 & 83.5 & 76.4 & 80.5 & 83.5 & 85.8 & \textcolor{RoyalBlue}{\textbf{84.8}} & \textcolor{RoyalBlue}{\textbf{72.6}} \\ \cline{2-14}  
 & \multirow{4}{*}{CL} & Base  & 27.7 & 32.4 & 37.2 & 37.6 & 33.0 & 29.2 & 32.0 & 32.2  & 32.3 & 27.8 & 20.7  \\
 &  & Base+SSF & \textcolor{RoyalBlue}{\textbf{32.1}} & \textcolor{RoyalBlue}{\textbf{37.3}} & 39.1 & 37.7 & 37.6 & 29.5 & 30.5 & 31.7 & 32.5 & \textcolor{RoyalBlue}{\textbf{30.9}} & \textcolor{RoyalBlue}{\textbf{21.9}}  \\ 
 &  & Base+SCF & 30.2 &38.0 &46.5 &46.7 &45.9 &41.3 &44.8 &43.7 &45.1 &34.3 &22.0\\
 &  & Ours (SF) & \textcolor{RoyalBlue}{\textbf{33.1}} & \textcolor{RoyalBlue}{\textbf{41.6}} & 46.4 & 47.6 & 46.7 & 41.2 & 44.7 & 43.3 & 45.6 & \textcolor{RoyalBlue}{\textbf{35.8}} & \textcolor{RoyalBlue}{\textbf{22.5}}  \\ 
 \bottomrule
 \end{tabular}
}
\label{tab:dataset_casiabn_seperate}
\end{table*}
We can see that SF surpasses the baseline on both conditions (NM + 4.3\%; BG + 7.1\%; CL + 9.0\%).
With the SF method, not only the accuracy of CL condition improved, but also the accuracy of NM, and BG, which means features from CL sequences can also provide useful information when recognizing a person, and they can not be neglected.
If the features before and after changing clothes are not correctly associated, the gait recognition model will miss important information, leading to insufficient learning and difficulty in distinguishing between different individuals.
However, due to the small dataset volume and lack of views in Outdoor-Gait, the upper bound with GaitGL backbone overfit\footnote{NM: 95.5, BG: 91.3, CL: 86.2 }, so we do not show the results with it.
\subsubsection{GREW}
To test the generalization of our method, we apply it to a large wild dataset GREW ~\cite{lin2014effects}. Our method excels in further separating the narrowed features within the feature space as comprehensively as possible. The results, as shown in Table~\ref{tab:grew}, illustrate that our proposed method outperforms the baseline under both conditions (Rank-1 + 1.9\%; Rank-5 + 1.6\%). This affirms that our Selective Fusion method can effectively guide the updating of clusters and finally get the clusters with different conditions. Additionally, it validates the generalization of our method in outdoor scenarios.
\begin{figure*}[t]
	\centering	 
	\includegraphics[width=\linewidth]{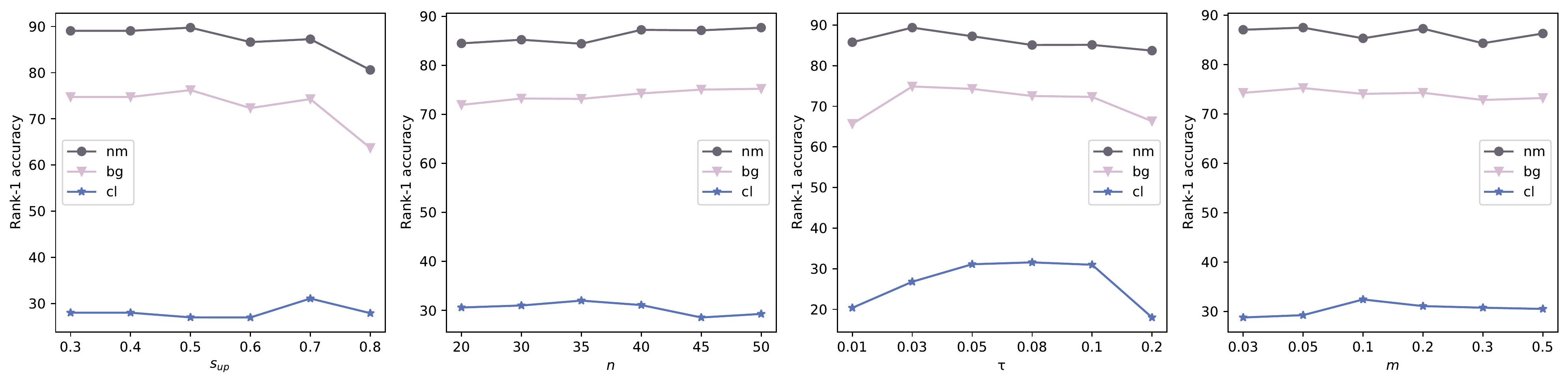}	 	
	\caption{The effect of hyper-parameters $s_{up}/n/\tau/m$ on baselines. In there we choose a set of hyper-parameters that have the best result in our experiments. Other hyper-parameters do not change the result a lot, just lead to sub-optimal.}
	\label{fig:hyperparameter}
\end{figure*}
\begin{figure*}[t]
	\centering	 
	\includegraphics[width=\linewidth]{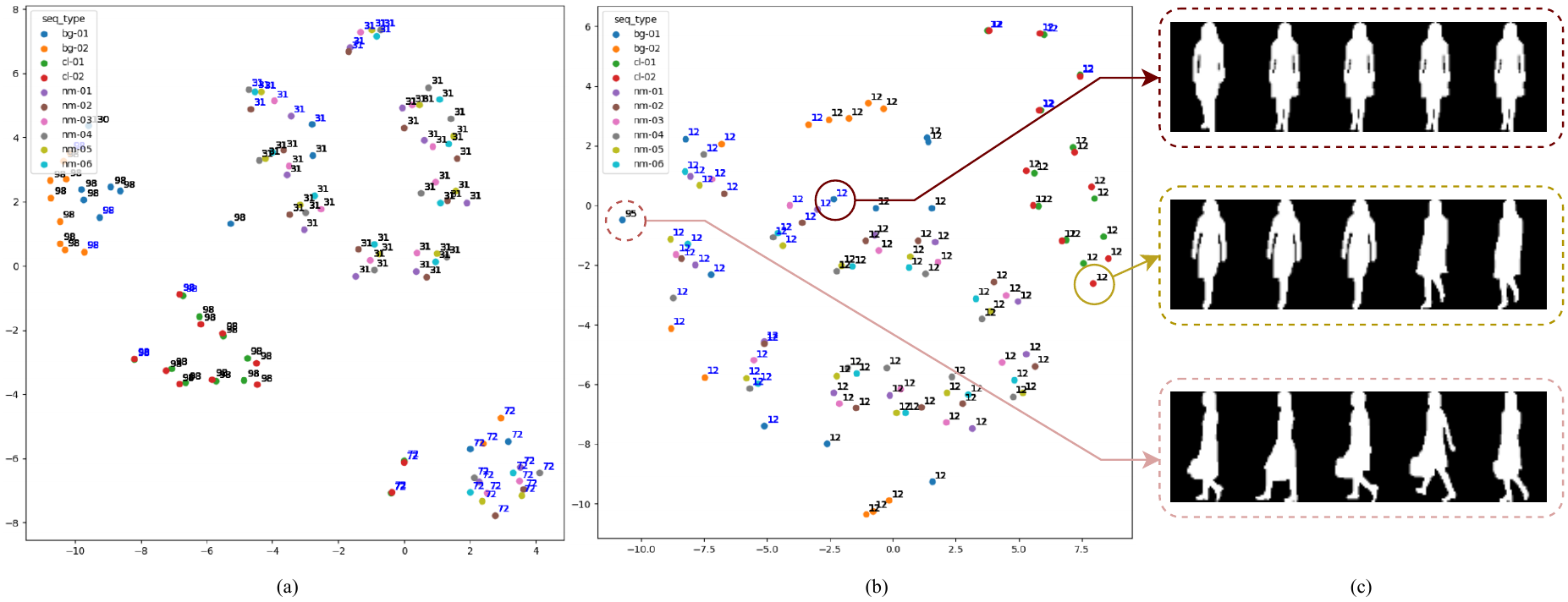}	 
	\caption{(a) and (b) are the TSNE images of the baseline method and our method(SF), respectively. The text above each feature point shows the pseudo label, and blue text indicates that the sequence is in front/back view. A color in a feature point represents a type of condition. Our method effectively clusters features belonging to the same person. (c) shows the corresponding gait sequence from (b), where the dashed circles indicate examples of clustering errors, and the solid circles indicate correct clustering examples. Please zoom in to see the details.}
	\label{fig:TSNE1}
\end{figure*}
\begin{table}[t]
\centering
\caption{The effect of cluster candidates number $k$ in support set.}
\setlength{\tabcolsep}{12pt}
\begin{tabular}{cccc}
\hline
Settings     & NM & BG & CL \\ \hline
Ours ($a=4$) & 89.2 & 81.2 & 35.1 \\
Ours ($a=3$) & 89.7 & 81.1 & 37.9   \\
Ours ($a=2$) & \textbf{90.3} & \textbf{82.9} & \textbf{40.8}  \\ \hline
\end{tabular}%
\label{tab:SC-Fusion}
\end{table}
\begin{table}[t]
\centering
\caption{The performance of different kinds of rates when conducting curriculum learning.}
\setlength{\tabcolsep}{12pt}
\begin{tabular}{cccc}
\toprule
Settings & NM & BG & CL \\ \midrule
Ours ($\lambda_{base}$) & 90.3  & 82.7 & 40.7 \\
Ours ($\lambda$) & \textbf{90.3} & \textbf{82.9} & \textbf{40.8} \\ \bottomrule
\end{tabular}%
\label{tab:SS-Fusion}
\end{table}
%
\begin{table}[h!]
\centering
\caption{Comparison of Rank-1 Accuracy (\%) between our method and GaitSSB* on the CASIA-BN and GREW datasets. GaitSSB* denotes the modified version of GaitSSB~\cite{fan2023learning} with an updated backbone.}
\label{tab:SF_gaitSSB}
\begin{tabular}{ccccc}
\hline
\multirow{2}{*}{Method} & \multicolumn{3}{c}{CASIA-BN}                                             & \multirow{2}{*}{GREW} \\
                         & NM            & BG            & CL                                 &                       \\ \hline
GaitSSB*                 & 43.0          & 30.2          & 31.1                               & 17.2                  \\
Ours                     & \textbf{90.3} & \textbf{82.9} & \multicolumn{1}{c}{\textbf{40.8}} & \textbf{20.2}         \\ \hline
\end{tabular}
\end{table}
\subsection{Ablation Study}
\subsubsection{Impact of Each Component in Our Selective Fusion}
In this section, we show that both SCF and SSF are essential components of our Selective Fusion method through experiments on the indoor CASIA-BN dataset and the outdoor datasets, Outdoor-Gait and GREW.
In Table~\ref{tab:each}, we show the results only using SCF or SSF.
Only with SSF, the rank-1 accuracy for each condition in front/back view slightly improved, but still had a poor performance on the cross-cloth problem.
When directly applying SCF, clusters mainly composed of sequences in front/back views will also pull closer to other clusters in the same condition, which should be forbidden.
Pulling more FVCs closer will further make these sequences merge into their actual clusters with other views, degrading the performance in recognizing sequences with front/back views.
Therefore, the best effect can be achieved only when these two methods are effectively combined.
As shown in Table~\ref{tab:dataset_casiabn_seperate}, we demonstrate the impact of SSF on sequences taken from front/back views, which validates its effectiveness. The improvements on sequences with front/back views are highlighted in blue.
%
\subsection{Effects of Different Parameters in Baseline}
\label{hyper}
Here we research how hyper-parameters $s_{up}$, $n$, $\tau$, $m$ affect the results of baselines. 
We adjust one parameter at the one time and keep other hyper-parameters unchanged. 
$s_{up}$ regulates the boundary of how far the features can be gathered into one cluster. The smaller $s_{up}$ it is, the tighter the boundary.
$n$ is the number of neighbors KNN searched for each sequence.
$\tau$ is the temperature parameter in ClusterNCE loss, indicating the entropy of the distribution.
$m$, the momentum value, controls the update speed of centroids stored in the Memory Bank.
From the results on CASIA-BN, we can see that when $s_{up}=0.7$, $n=40$, $\tau=0.05$, $m=0.2$ we have the overall best results for NM, BG, and CL.
When these parameters deviate too much from the current setting, the performance is sub-optimal.
Here we show the accuracy of NM, BG, and CL when adopting different parameters in the baseline framework in Figure~\ref{fig:hyperparameter}.
\subsubsection{Impact of Candidate Number in Support Set}
Here we discuss the effect of a parameter in SCF, the candidate number $a$, in the support set.
In Table~\ref{tab:SC-Fusion}, we can see that when $a=2$, we have the best performance, which is in line with the fact that the features of NM and BG are easily projected together in the feature space since they have larger similarity and we should drag the features of NM with CL specifically in CASIA-BN.
\subsubsection{Impact of Rate of Curriculum Learning in SSF}
We test the effect of SSF with a dynamic or constant rate when conducting curriculum learning on CASIA-BN. 
Without curriculum learning, linearly clustering the front/back view sequences with sequences in other views will degrade the performance.
In Table~\ref{tab:SS-Fusion}, with a dynamic pulling rate, we can relax the requirement when training the model, which can make the model learn from easy to hard better. 
\subsubsection{Instance vs. Cluster-based method}
To further validate the effectiveness of our method, we conducted extensive experiments on both the indoor CASIA-BN dataset and the outdoor GREW dataset, comparing our approach against GaitSSB*. For a fair comparison, we replaced the backbone of GaitSSB with GaitSet while keeping all other experimental settings unchanged. As shown in Table~\ref{tab:SF_gaitSSB}, our method consistently outperforms GaitSSB* on both datasets. GaitSSB is an instance-based contrastive learning method that constructs positive pairs at the instance level through data augmentation. However, due to the limitations of current data augmentation techniques for gait sequences, and more importantly, the positive pairs are drawn from the same sequence, resulting in positive sample pairs that are very similar to each other. This makes it difficult to simulate the variations caused by changes in clothing and camera viewpoints, limiting the ability to provide effective supervisory signals that can guide the model to learn robust features. 

In contrast, our method adopts a clustering strategy to group unlabeled data and generate pseudo-labels, allowing the model to learn representations based on cluster assignments. By integrating clustering strategies with selective fusion, our approach effectively addresses the practical challenges of unsupervised gait recognition. As shown in Table 9, the experimental results demonstrate that our method is more robust in handling complex scenarios, such as cross-clothing and cross-view variations.

\subsubsection{Impact of different training dataset scales}
We further conducted related experiments on CASIA-BN to verify the impact of different training dataset scales on our method in Table~\ref{tab:train_scale}. The CASIA-BN training set consists of 8,107 sequences. We randomly selected different training dataset scales (4,000, 6,000, and all) from the CASIA-BN training set to train our method. The performance of the models trained on different scales was evaluated using the standard test set. As shown in Table~\ref{tab:train_scale}, having more training data is beneficial, as a larger dataset enables the model to learn more generalizable features, ultimately improving recognition performance.

\begin{table}[h]
\centering
\caption{Randomly selecting different training dataset scales from the CASIA-BN training set to train our method, and evaluating the Rank-1 accuracy (\%)on the standard test set.}
\label{tab:train_scale}
\begin{tabular}{cccc}
\hline
\multirow{2}{*}{Training dataset Scale} & \multicolumn{3}{c}{CASIA-BN} \\
                                        & NM       & BG      & CL      \\ \hline
4000                                    & 86.9     & 75.8    & 33.6    \\
6000                                    & 89.0     & 79.1    & 37.7    \\
All                                     & 90.3     & 82.9    & 40.8    \\ \hline
\end{tabular}
\end{table}

\subsubsection{Visulization of Selective Fusion}
The visualization effect of Selective Fusion is shown in Figure~\ref{fig:TSNE1}. We selected a subject in CASIA-BN and found that in baseline, BG and CL have a different pseudo label from NM. 
At the same time, some sequences in front/back views of NM/BG/CL are also assigned different pseudo labels with sequences in other views.
With SF, as shown in Figure~\ref{fig:TSNE1}(b), most sequences from various views and conditions are assigned the same identity. As shown in Figure~\ref{fig:TSNE1}(c), we also visualized some cases of SF.
\section{Limitation and future work}
Our method can be employed with off-the-shelf backbones to train on a new, unlabeled dataset. However, there are still some limitations:
First, our approach relies on prior knowledge in certain areas. For instance, (1) we use knowledge of viewing angles to identify challenging samples with front/back views, and (2) our model requires well-initialized parameters to ensure reasonable clustering performance at the beginning of training. These dependencies on prior knowledge can impact the final accuracy. To overcome this, we need to explore methods that can achieve better unsupervised gait recognition without such reliance.

Additionally, data augmentation plays a crucial role in unsupervised gait recognition. However, our current augmentation method only simulates clothing variations in specific scenarios. To better capture real-world variations in gait sequences, it is necessary to develop more robust data augmentation techniques.

\section{Conclusion}
In this work, we propose a new task, Unsupervised Gait Recognition. 
We first design a new baseline with cluster-level contrastive learning.
We identified two key challenges in unsupervised gait recognition: (1) sequences with different clothing are not grouped into a single cluster, and (2) sequences captured from front/back views are difficult to merge with those from other views.
To address these challenges, we developed the Selective Fusion method, which includes Selective Cluster Fusion and Selective Sample Fusion. Selective Cluster Fusion helps cluster sequences of the same individual across different clothing conditions, while Selective Sample Fusion progressively merges sequences from front/back views with those from other views.

Our experiments demonstrate that the proposed method effectively improves accuracy under both clothing variation and front/back view conditions.
This work reduces reliance on labeled data, enabling us to learn high-quality feature representations from large-scale unlabeled datasets, thereby advancing the development of gait recognition.

\ifCLASSOPTIONcaptionsoff
  \newpage
\fi

\bibliographystyle{IEEEtran}
\bibliography{Transactions-Bibliography/IEEEabrv,Transactions-Bibliography/egbib}

\begin{thebibliography}{10}
\providecommand{\url}[1]{#1}
\csname url@samestyle\endcsname
\providecommand{\newblock}{\relax}
\providecommand{\bibinfo}[2]{#2}
\providecommand{\BIBentrySTDinterwordspacing}{\spaceskip=0pt\relax}
\providecommand{\BIBentryALTinterwordstretchfactor}{4}
\providecommand{\BIBentryALTinterwordspacing}{\spaceskip=\fontdimen2\font plus
\BIBentryALTinterwordstretchfactor\fontdimen3\font minus \fontdimen4\font\relax}
\providecommand{\BIBforeignlanguage}[2]{{%
\expandafter\ifx\csname l@#1\endcsname\relax
\typeout{** WARNING: IEEEtran.bst: No hyphenation pattern has been}%
\typeout{** loaded for the language `#1'. Using the pattern for}%
\typeout{** the default language instead.}%
\else
\language=\csname l@#1\endcsname
\fi
#2}}
\providecommand{\BIBdecl}{\relax}
\BIBdecl

\bibitem{chao2019gaitset}
H.~Chao, Y.~He, J.~Zhang, and J.~Feng, ``Gaitset: Regarding gait as a set for cross-view gait recognition,'' in \emph{Proceedings of the AAAI conference on artificial intelligence}, vol.~33, no.~01, 2019, pp. 8126--8133.

\bibitem{fan2020gaitpart}
C.~Fan, Y.~Peng, C.~Cao, X.~Liu, S.~Hou, J.~Chi, Y.~Huang, Q.~Li, and Z.~He, ``Gaitpart: Temporal part-based model for gait recognition,'' in \emph{Proceedings of the IEEE/CVF conference on computer vision and pattern recognition}, 2020, pp. 14\,225--14\,233.

\bibitem{lin2021gait}
B.~Lin, S.~Zhang, and X.~Yu, ``Gait recognition via effective global-local feature representation and local temporal aggregation,'' in \emph{Proceedings of the IEEE/CVF International Conference on Computer Vision}, 2021, pp. 14\,648--14\,656.

\bibitem{ren2022progressive}
X.~Ren, S.~Hou, C.~Cao, X.~Liu, and Y.~Huang, ``Progressive feature learning for realistic cloth-changing gait recognition,'' \emph{arXiv preprint arXiv:2207.11720}, 2022.

\bibitem{yu2006framework}
S.~Yu, D.~Tan, and T.~Tan, ``A framework for evaluating the effect of view angle, clothing and carrying condition on gait recognition,'' in \emph{18th international conference on pattern recognition (ICPR'06)}, vol.~4.\hskip 1em plus 0.5em minus 0.4em\relax IEEE, 2006, pp. 441--444.

\bibitem{lin2019bottom}
Y.~Lin, X.~Dong, L.~Zheng, Y.~Yan, and Y.~Yang, ``A bottom-up clustering approach to unsupervised person re-identification,'' in \emph{Proceedings of the AAAI conference on artificial intelligence}, vol.~33, no.~01, 2019, pp. 8738--8745.

\bibitem{ge2020self}
Y.~Ge, F.~Zhu, D.~Chen, R.~Zhao \emph{et~al.}, ``Self-paced contrastive learning with hybrid memory for domain adaptive object re-id,'' \emph{Advances in Neural Information Processing Systems}, vol.~33, pp. 11\,309--11\,321, 2020.

\bibitem{lin2020unsupervised}
Y.~Lin, L.~Xie, Y.~Wu, C.~Yan, and Q.~Tian, ``Unsupervised person re-identification via softened similarity learning,'' in \emph{Proceedings of the IEEE/CVF conference on computer vision and pattern recognition}, 2020, pp. 3390--3399.

\bibitem{wang2020unsupervised}
D.~Wang and S.~Zhang, ``Unsupervised person re-identification via multi-label classification,'' in \emph{Proceedings of the IEEE/CVF conference on computer vision and pattern recognition}, 2020, pp. 10\,981--10\,990.

\bibitem{dai2021cluster}
Z.~Dai, G.~Wang, W.~Yuan, S.~Zhu, and P.~Tan, ``Cluster contrast for unsupervised person re-identification,'' in \emph{Proceedings of the Asian Conference on Computer Vision}, 2022, pp. 1142--1160.

\bibitem{9978648}
K.~Nikhal and B.~S. Riggan, ``Multi-context grouped attention for unsupervised person re-identification,'' \emph{IEEE Transactions on Biometrics, Behavior, and Identity Science}, pp. 1--1, 2022.

\bibitem{ball2012unsupervised}
A.~Ball, D.~Rye, F.~Ramos, and M.~Velonaki, ``Unsupervised clustering of people from'skeleton'data,'' in \emph{Proceedings of the seventh annual ACM/IEEE international conference on Human-Robot Interaction}, 2012, pp. 225--226.

\bibitem{cola2015unsupervised}
G.~Cola, M.~Avvenuti, A.~Vecchio, G.-Z. Yang, and B.~Lo, ``An unsupervised approach for gait-based authentication,'' in \emph{2015 IEEE 12th International Conference on Wearable and Implantable Body Sensor Networks (BSN)}.\hskip 1em plus 0.5em minus 0.4em\relax IEEE, 2015, pp. 1--6.

\bibitem{rida2015unsupervised}
I.~Rida, S.~Al~Maadeed, and A.~Bouridane, ``Unsupervised feature selection method for improved human gait recognition,'' in \emph{2015 23rd European Signal Processing Conference (EUSIPCO)}.\hskip 1em plus 0.5em minus 0.4em\relax IEEE, 2015, pp. 1128--1132.

\bibitem{zhang2024research}
J.~Zhang, J.~Cao, J.~Chang, X.~Li, H.~Liu, and Z.~Li, ``Research on the application of computer vision based on deep learning in autonomous driving technology,'' \emph{arXiv preprint arXiv:2406.00490}, 2024.

\bibitem{liao2020model}
R.~Liao, S.~Yu, W.~An, and Y.~Huang, ``A model-based gait recognition method with body pose and human prior knowledge,'' \emph{Pattern recognition}, vol.~98, p. 107069, 2020.

\bibitem{li2020jointsgait}
N.~Li, X.~Zhao, and C.~Ma, ``Jointsgait: A model-based gait recognition method based on gait graph convolutional networks and joints relationship pyramid mapping,'' \emph{arXiv preprint arXiv:2005.08625}, 2020.

\bibitem{9916067}
J.~Chen, Z.~Wang, P.~Yi, K.~Zeng, Z.~He, and Q.~Zou, ``Gait pyramid attention network: Toward silhouette semantic relation learning for gait recognition,'' \emph{IEEE Transactions on Biometrics, Behavior, and Identity Science}, vol.~4, no.~4, pp. 582--595, 2022.

\bibitem{9229117}
A.~Sepas-Moghaddam and A.~Etemad, ``View-invariant gait recognition with attentive recurrent learning of partial representations,'' \emph{IEEE Transactions on Biometrics, Behavior, and Identity Science}, vol.~3, no.~1, pp. 124--137, 2021.

\bibitem{9913216}
X.~Huang, X.~Wang, B.~He, S.~He, W.~Liu, and B.~Feng, ``Star: Spatio-temporal augmented relation network for gait recognition,'' \emph{IEEE Transactions on Biometrics, Behavior, and Identity Science}, vol.~5, no.~1, pp. 115--125, 2023.

\bibitem{10042966}
R.~Wang, Y.~Shi, H.~Ling, Z.~Li, P.~Li, B.~Liu, H.~Zheng, and Q.~Wang, ``Gait recognition via gait period set,'' \emph{IEEE Transactions on Biometrics, Behavior, and Identity Science}, pp. 1--1, 2023.

\bibitem{hou2022gait}
S.~Hou, X.~Liu, C.~Cao, and Y.~Huang, ``Gait quality aware network: Toward the interpretability of silhouette-based gait recognition,'' \emph{{IEEE} Trans. Neural Netw.}, 2022.

\bibitem{das2023gait}
D.~Das, A.~Agarwal, and P.~Chattopadhyay, ``Gait recognition from occluded sequences in surveillance sites,'' in \emph{Computer Vision--ECCV 2022 Workshops: Tel Aviv, Israel, October 23--27, 2022, Proceedings, Part V}.\hskip 1em plus 0.5em minus 0.4em\relax Springer, 2023, pp. 703--719.

\bibitem{9870842}
S.~Zhang, Y.~Wang, and A.~Li, ``Gait energy image-based human attribute recognition using two-branch deep convolutional neural network,'' \emph{IEEE Transactions on Biometrics, Behavior, and Identity Science}, vol.~5, no.~1, pp. 53--63, 2023.

\bibitem{9928336}
S.~Hou, C.~Fan, C.~Cao, X.~Liu, and Y.~Huang, ``A comprehensive study on the evaluation of silhouette-based gait recognition,'' \emph{IEEE Transactions on Biometrics, Behavior, and Identity Science}, pp. 1--1, 2022.

\bibitem{yan2018spatial}
S.~Yan, Y.~Xiong, and D.~Lin, ``Spatial temporal graph convolutional networks for skeleton-based action recognition,'' in \emph{Proceedings of the AAAI conference on artificial intelligence}, 2018.

\bibitem{gaitgraph}
T.~Torben, K.~Ali, G.~Johannes, H.~Fabian, and H.~Stefan, ``Gaitgraph: Graph convolutional network for skeleton-based gait recognition,'' in \emph{IEEE International Conference on Image Processing (ICIP)}.\hskip 1em plus 0.5em minus 0.4em\relax IEEE, 2021, p. 2314–2318.

\bibitem{chen2020simple}
T.~Chen, S.~Kornblith, M.~Norouzi, and G.~Hinton, ``A simple framework for contrastive learning of visual representations,'' in \emph{International conference on machine learning}.\hskip 1em plus 0.5em minus 0.4em\relax PMLR, 2020, pp. 1597--1607.

\bibitem{chen2021exploring}
X.~Chen and K.~He, ``Exploring simple siamese representation learning,'' in \emph{Proceedings of the IEEE/CVF conference on computer vision and pattern recognition}, 2021, pp. 15\,750--15\,758.

\bibitem{he2020momentum}
K.~He, H.~Fan, Y.~Wu, S.~Xie, and R.~Girshick, ``Momentum contrast for unsupervised visual representation learning,'' in \emph{Proceedings of the IEEE/CVF conference on computer vision and pattern recognition}, 2020, pp. 9729--9738.

\bibitem{fan2023learning}
C.~Fan, S.~Hou, J.~Wang, Y.~Huang, and S.~Yu, ``Learning gait representation from massive unlabelled walking videos: A benchmark,'' \emph{IEEE Transactions on Pattern Analysis and Machine Intelligence}, 2023.

\bibitem{zeng2020hierarchical}
K.~Zeng, M.~Ning, Y.~Wang, and Y.~Guo, ``Hierarchical clustering with hard-batch triplet loss for person re-identification,'' in \emph{Proceedings of the IEEE/CVF conference on computer vision and pattern recognition}, 2020, pp. 13\,657--13\,665.

\bibitem{wang2021camera}
M.~Wang, B.~Lai, J.~Huang, X.~Gong, and X.-S. Hua, ``Camera-aware proxies for unsupervised person re-identification,'' in \emph{Proceedings of the AAAI conference on artificial intelligence}, vol.~35, no.~4, 2021, pp. 2764--2772.

\bibitem{chen2021ice}
H.~Chen, B.~Lagadec, and F.~Bremond, ``Ice: Inter-instance contrastive encoding for unsupervised person re-identification,'' in \emph{Proceedings of the IEEE/CVF International Conference on Computer Vision}, 2021, pp. 14\,960--14\,969.

\bibitem{xuan2021intra}
S.~Xuan and S.~Zhang, ``Intra-inter camera similarity for unsupervised person re-identification,'' in \emph{Proceedings of the IEEE/CVF conference on computer vision and pattern recognition}, 2021, pp. 11\,926--11\,935.

\bibitem{zhang2023camera}
G.~Zhang, H.~Zhang, W.~Lin, A.~K. Chandran, and X.~Jing, ``Camera contrast learning for unsupervised person re-identification,'' \emph{{IEEE} Trans. Circuits Syst. Video Technol.}, 2023.

\bibitem{chen2023class}
Z.~Chen, L.~Jing, L.~Yang, Y.~Li, and B.~Li, ``Class-level confidence based 3d semi-supervised learning,'' in \emph{Proceedings of the IEEE/CVF Winter Conference on Applications of Computer Vision}, 2023, pp. 633--642.

\bibitem{chen2021multimodal}
Z.~Chen, L.~Jing, Y.~Liang, Y.~Tian, and B.~Li, ``Multimodal semi-supervised learning for 3d objects,'' \emph{arXiv preprint arXiv:2110.11601}, 2021.

\bibitem{li2022unsupervised}
M.~Li, P.~Xu, X.~Zhu, and J.~Guo, ``Unsupervised long-term person re-identification with clothes change,'' \emph{arXiv preprint arXiv:2202.03087}, 2022.

\bibitem{bengio2009curriculum}
Y.~Bengio, J.~Louradour, R.~Collobert, and J.~Weston, ``Curriculum learning,'' in \emph{Proceedings of the 26th annual international conference on machine learning}, 2009, pp. 41--48.

\bibitem{takemura2018multi}
N.~Takemura, Y.~Makihara, D.~Muramatsu, T.~Echigo, and Y.~Yagi, ``Multi-view large population gait dataset and its performance evaluation for cross-view gait recognition,'' \emph{IPSJ Transactions on Computer Vision and Applications}, vol.~10, no.~1, pp. 1--14, 2018.

\bibitem{fix1989discriminatory}
E.~Fix and J.~L. Hodges, ``Discriminatory analysis. nonparametric discrimination: Consistency properties,'' \emph{International Statistical Review/Revue Internationale de Statistique}, vol.~57, no.~3, pp. 238--247, 1989.

\bibitem{rosvall2008maps}
M.~Rosvall and C.~T. Bergstrom, ``Maps of random walks on complex networks reveal community structure,'' \emph{Proceedings of the national academy of sciences}, vol. 105, no.~4, pp. 1118--1123, 2008.

\bibitem{song2019gaitnet}
C.~Song, Y.~Huang, Y.~Huang, N.~Jia, and L.~Wang, ``Gaitnet: An end-to-end network for gait based human identification,'' \emph{Pattern recognition}, vol.~96, p. 106988, 2019.

\bibitem{hou2019learning}
S.~Hou, X.~Pan, C.~C. Loy, Z.~Wang, and D.~Lin, ``Learning a unified classifier incrementally via rebalancing,'' in \emph{Proceedings of the IEEE/CVF conference on computer vision and pattern recognition}, 2019, pp. 831--839.

\bibitem{paszke2019pytorch}
A.~Paszke, S.~Gross, F.~Massa, A.~Lerer, J.~Bradbury, G.~Chanan, T.~Killeen, Z.~Lin, N.~Gimelshein, L.~Antiga \emph{et~al.}, ``Pytorch: An imperative style, high-performance deep learning library,'' \emph{Advances in Neural Information Processing Systems}, vol.~32, pp. 8026--8037, 2019.

\bibitem{lin2014effects}
C.-C. Lin, K.-C. Huang \emph{et~al.}, ``Effects of lighting color, illumination intensity, and text color on visual performance,'' \emph{International Journal of Applied Science and Engineering}, vol.~12, no.~3, pp. 193--202, 2014.

\bibitem{he2016deep}
K.~He, X.~Zhang, S.~Ren, and J.~Sun, ``Deep residual learning for image recognition,'' in \emph{Proceedings of the IEEE/CVF conference on computer vision and pattern recognition}, 2016, pp. 770--778.

\bibitem{deng2009imagenet}
J.~Deng, W.~Dong, R.~Socher, L.-J. Li, K.~Li, and L.~Fei-Fei, ``Imagenet: A large-scale hierarchical image database,'' in \emph{Proceedings of the IEEE/CVF conference on computer vision and pattern recognition}.\hskip 1em plus 0.5em minus 0.4em\relax Ieee, 2009, pp. 248--255.

\end{thebibliography}



%

%
\section{Biography Section}

\begin{IEEEbiography}[{\includegraphics[width=1in,height=1.25in,clip,keepaspectratio]{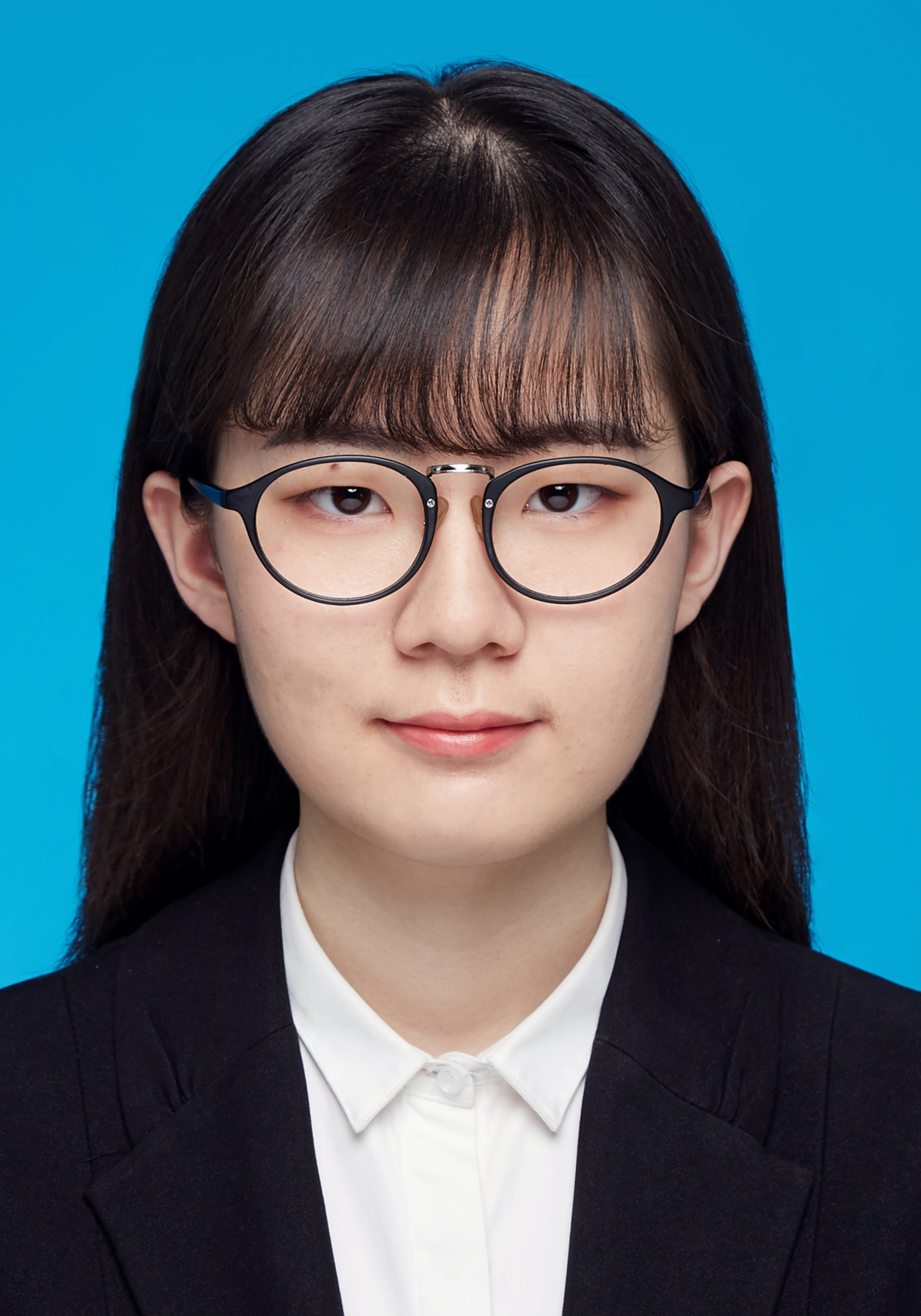}}]{Xuqian Ren}
received the B.E. degree from the University of Science and Technology Beijing in 2019, and received the M.S. degree from Beijing Institute of Technology in 2022. She is currently a Ph.D. candidate in the Computer Science, Faculty of Information Technology and Communication Sciences, at Tampere University, Finland. Her current research interests include Novel view synthesis, image generation, and 3d reconstruction.
\end{IEEEbiography}

\vspace{-10mm}
\begin{IEEEbiography}[{\includegraphics[width=1in,height=1.25in,clip,keepaspectratio]{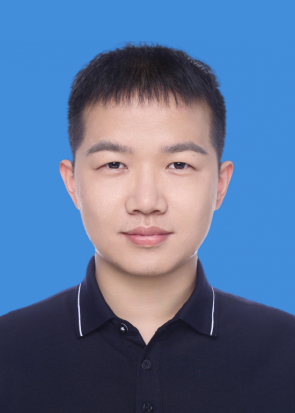}}]{Shaopeng Yang}
received the B.E. degree from Langfang Normal University in 2015, received the M.S. degree from Beijing Union University in 2019. He is currently a Ph.D. student with School of Artificial Intelligence, Beijing Normal University. He current research interests include contrastive learning and gait recognition.
\end{IEEEbiography}

\vspace{-10mm}
\begin{IEEEbiography}[{\includegraphics[width=1in,height=1.25in,clip,keepaspectratio]{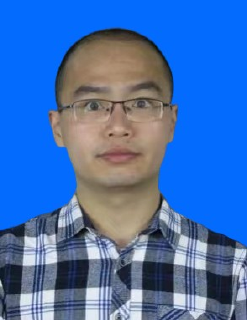}}]{Saihui Hou}
received the B.E. and Ph.D. degrees from University of Science and Technology of China in 2014 and 2019, respectively. He is currently an Assistant Professor with School of Artificial Intelligence, Beijing Normal University, and works in cooperation with Watrix Technology Limited Co. Ltd. His research interests include computer vision and machine learning. He recently focuses on gait recognition which aims to identify different people according to the walking patterns.
\end{IEEEbiography}

\vspace{-10mm}
\begin{IEEEbiography}[{\includegraphics[width=1in,height=1.25in,clip,keepaspectratio]{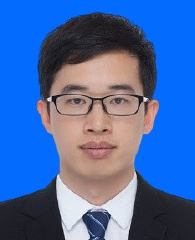}}]{Chunshui Cao}
received the B.E. and Ph.D. degrees from University of Science and Technology of China in 2013 and 2018, respectively. During his Ph.D. study, he joined Center for Research on Intelligent Perception and Computing, National Laboratory of Pattern Recognition, Institute of Automation, Chinese Academy of Sciences. From 2018 to 2020, he worked as a Postdoctoral Fellow with PBC School of Finance, Tsinghua University. He is currently a Research Scientist with Watrix Technology Limited Co. Ltd. His research interests include pattern recognition, computer vision and machine learning.
\end{IEEEbiography}

\vspace{-10mm}
\begin{IEEEbiography}[{\includegraphics[width=1in,height=1.25in,clip,keepaspectratio]{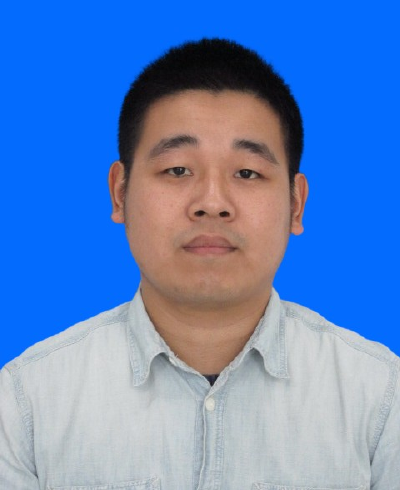}}]{Xu Liu}
received the B.E. and Ph.D. degrees from University of Science and Technology of China in 2013 and 2018, respectively. He is currently a Research Scientist with Watrix Technology Limited Co. Ltd. His research interests include gait recognition, object detection and image segmentation.
\end{IEEEbiography}

\vspace{-10mm}
\begin{IEEEbiography}[{\includegraphics[width=1in,height=1.25in,clip,keepaspectratio]{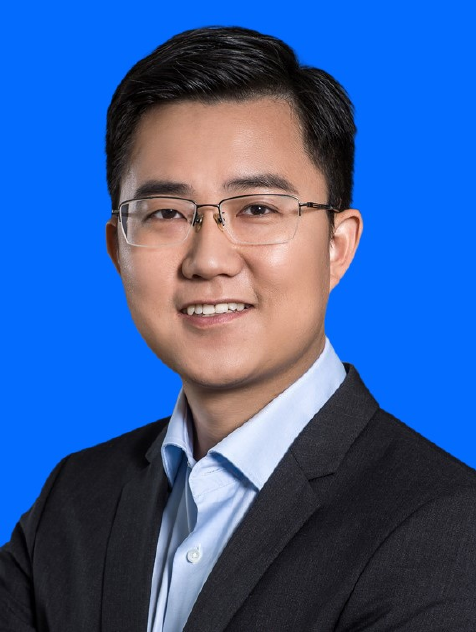}}]{Yongzhen Huang}
received the B.E. degree from Huazhong University of Science and Technology in 2006, and the Ph.D. degree from Institute of Automation, Chinese Academy of Sciences in 2011. He is currently a Professor with School of Artificial Intelligence, Beijing Normal University, and works in cooperation with Watrix Technology Limited Co. Ltd. He has published one book and more than 80 papers at international journals and conferences such as TPAMI, IJCV, TIP, TSMCB, TMM, TCSVT, CVPR, ICCV, ECCV, NIPS, AAAI. His research interests include pattern recognition, computer vision and machine learning.
\end{IEEEbiography}



\end{document}


\setcounter{table}{7}  
\setcounter{figure}{4}  
%
\title{Supplementary Materials of Unsupervised \\ Gait Recognition with Selective Fusion}

\author{Xuqian Ren,
        Shaopeng Yang,
        Saihui Hou,
        Chunshui Cao,
        Xu Liu and
        Yongzhen Huang
}

\markboth{Journal of \LaTeX\ Class Files,~Vol.~14, No.~8, August~2015}%
{Shell \MakeLowercase{\textit{et al.}}: Bare Demo of IEEEtran.cls for IEEE Journals}
%



\maketitle

\textcolor{blue}{
In the supplementary materials, we provide implementation details, ablation studies, and visualizations of t-SNE to further illustrate the effectiveness of our method.}
\section{Implementation Details}
\textcolor{blue}{For CASIA-BN~\cite{yu2006framework} and Outdoor-Gait~\cite{song2019gaitnet}, we achieve a good performance by simply using a fixed momentum value of $m=0.2$. However, due to its unprecedented scale and real-world settings, the utilization of a relatively small fixed value for GREW~\cite{lin2014effects} will result in excessively rapid updates of cluster centroids, compromising the overall consistency of the clusters. Therefore, we use a cosine annealing strategy to update $m$, the new strategy can be formulated as follows:
\begin{equation}
\label{momentum_strategy}
   m_t = m_{\text{min}} + \frac{1}{2} (m_{\text{max}} - m_{\text{min}}) \left(1 + \cos\left(\frac{t \pi}{T}\right)\right)
\end{equation}
where $m_t$ is the momentum at training step $t$, $m_{\text{min}}$ is the minimum momentum value, $m_{\text{max}}$ is the initial value, and $T$ represents the total number of training steps within a single epoch. As training progresses ($t$ increases), the momentum decreases following a cosine curve. We set $m_{\text{max}}=0.5$ and $m_{\text{min}}=0.1$.
}
\section{Ablation Study}
\subsection{Effects of Different Parameters in Baseline}
\label{hyper}
\begin{figure*}[h]
	\centering	 
	\includegraphics[width=\linewidth]{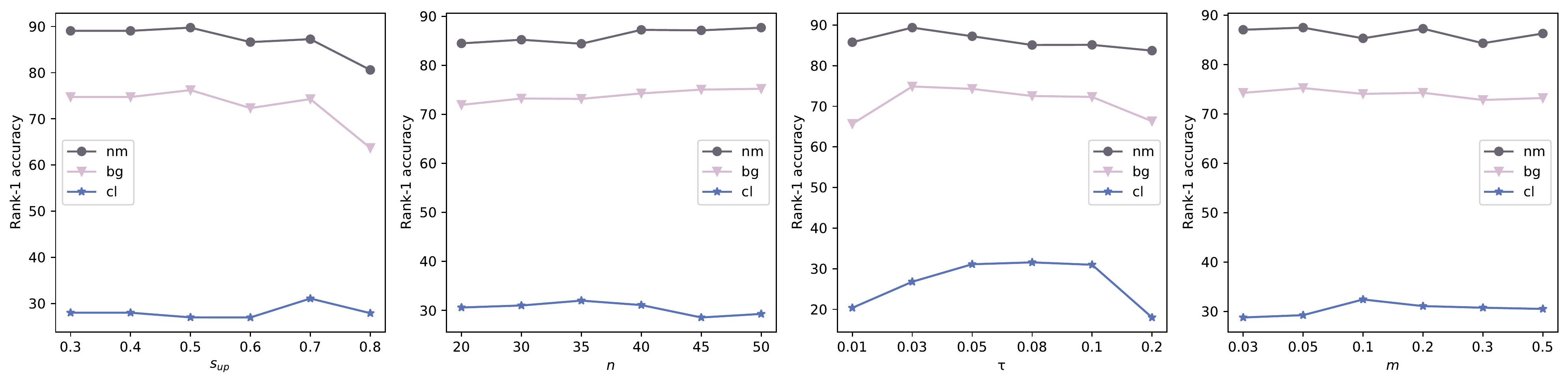}	 	
	\caption{The effect of hyper-parameters $s_{up}/n/\tau/m$ on baselines. In there we choose a set of hyper-parameters that have the best result in our experiments. Other hyper-parameters don't change the result a lot, just lead to sub-optimal.}
	\label{fig:hyperparameter}
\end{figure*}

Here we research how hyper-parameters $s_{up}$, $n$, $\tau$, $m$ affect the results of baselines. 
We adjust one parameter at the one time and keep other hyper-parameters unchanged. 
$s_{up}$ regulates the boundary of how far the features can be gathered into one cluster. The smaller $s_{up}$ it is, the tighter the boundary.
$n$ is the number of neighbors KNN searched for each sequence.
$\tau$ is the temperature parameter in ClusterNCE loss, indicating the entropy of the distribution.
$m$, the momentum value, controls the update speed of centroids stored in the Memory Bank.
From the results on CASIA-BN, we can see that when $s_{up}=0.7$, $n=40$, $\tau=0.05$, $m=0.2$ we have the overall best results for NM, BG, and CL.
When these parameters deviate too much from the current setting, the performance is sub-optimal.
Here we show the accuracy of NM, BG, and CL when adopting different parameters in the baseline framework in Figure~\ref{fig:hyperparameter}.

\subsection{Impact of Rate of Curriculum Learning in SSF}

We test the effect of SSF with a dynamic or constant rate when conducting curriculum learning on CASIA-BN. 
Without curriculum learning, linearly clustering the front/back view sequences with sequences in other views will degrade the performance.
In Table~\ref{tab:SS-Fusion}, with a dynamic pulling rate, we can relax the requirement when training the model, which can make the model learn from easy to hard better. 
\begin{table}[t]
\centering
\caption{The performance of different kinds of rates when conducting curriculum learning.}
\setlength{\tabcolsep}{12pt}
\begin{tabular}{cccc}
\toprule
Settings & NM & BG & CL \\ \midrule
Ours ($\lambda_{base}$) & 90.3  & 82.7 & 40.7 \\
Ours ($\lambda$) & \textbf{90.3} & \textbf{82.9} & \textbf{40.8} \\ \bottomrule
\end{tabular}%
\label{tab:SS-Fusion}
\end{table}

\section{Visulization of Selective Fusion}
\begin{figure*}[t]
\centering  
\subfigure[Baseline]{
\label{Fig.sub.1}
\includegraphics[width=0.3\linewidth]{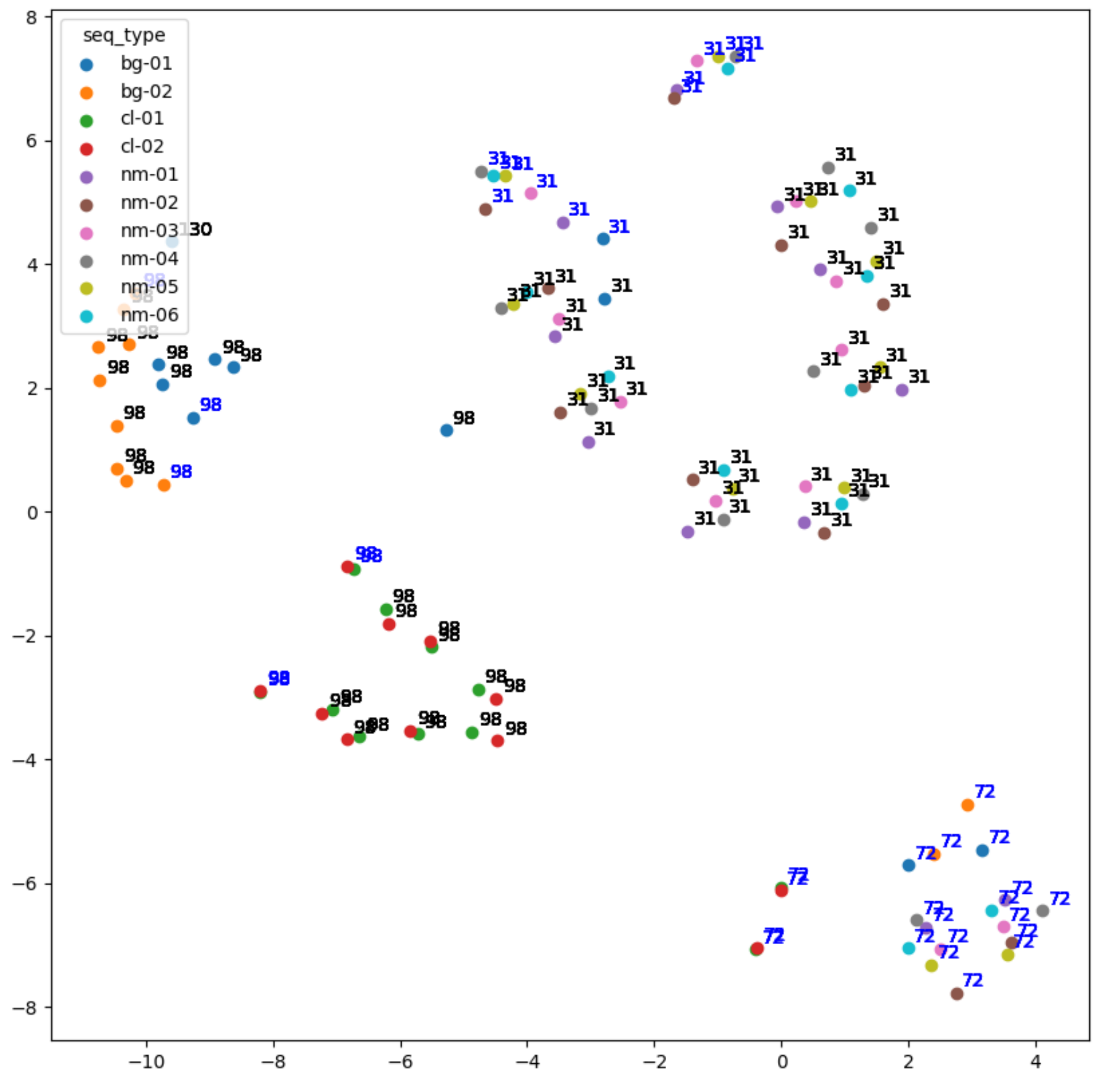}}\subfigure[Ours]{
\label{Fig.sub.2}
\includegraphics[width=0.3\linewidth]{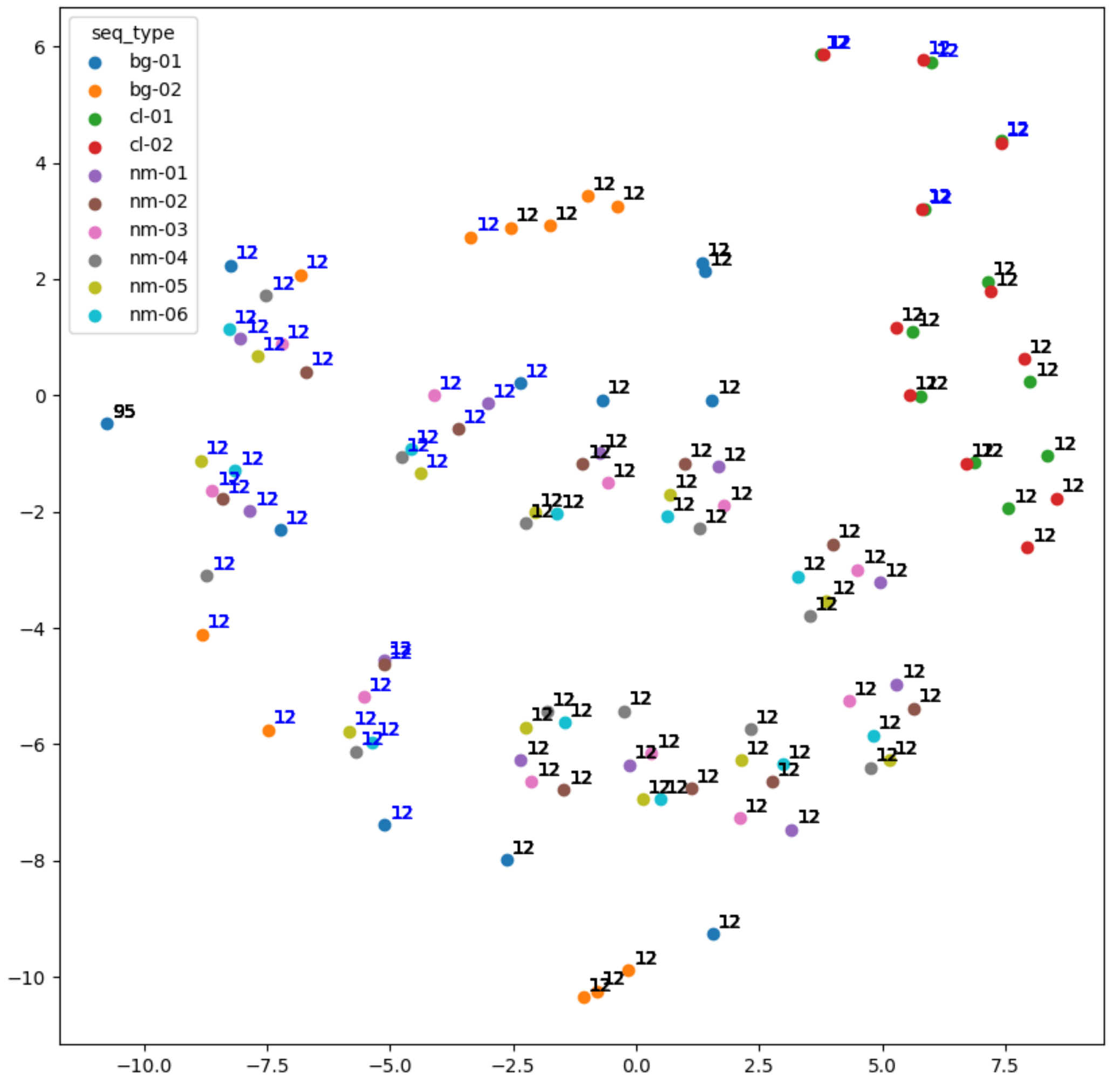}}
\caption{The TSNE images of baseline/Ours. The text above each feature point shows the pseudo label. The blue color in the text represents \textit{the sequence is in front/back view}. A type of color in a feature point indicates a type of condition. SF is the Selective Fusion method we proposed. Please scroll in to see the details.}
\label{fig:tsne}
\end{figure*}

The visualization effect of Selective Fusion is shown in Figure~\ref{fig:tsne}. We select a subject in CASIA-BN, finding that in baseline, BG and CL have a different pseudo label with NM. 
At the same time, some sequences in front/back views of NM/BG/CL are also assigned different pseudo labels with sequences in other views.
With Selective Fusion, most sequences in various views and different conditions are assigned the same ID.
\section{Discussion and future work}

Our method can be employed with off-the-shelf backbones to train a new unlabeled dataset without a label. 
However, there are still some limitations. 
First, we utilize prior knowledge about the views to identify the challenging samples with front/back views. Without such knowledge, it is difficult to specifically group these sequences with other views.
More automatic methods can be further developed to reduce the dependence on prior knowledge.

Second, our method uses data augmentation to simulate cross-cloth samples for gait recognition, which makes the model's recognition performance dependent on the effect of data augmentation to some extent. Therefore, more data augmentation methods can be developed to simulate real-world cloth-changing situations.
Also, the intrinsic principles of cross-cloth recognition need further study.

Third, the accuracy of unsupervised learning relies much on the pre-trained model's accuracy, so a higher precision model is necessary when conducting unsupervised learning on larger datasets, especially on real-world datasets.

Since OU-MVLP~\cite{takemura2018multi} is a dataset that is captured in the lab environment, it provides limited knowledge when realizing cross-view and cross-cloth on real-world data.
So further efforts need to be spent on training a more robust and high precision pre-train model to make the unsupervised learning method better adapted to the real world. 

Our work conducts gait recognition with unsupervised learning, which alleviates the human labor requirement in the data collection process, making training gait recognition models with large datasets possible and economical. This is crucial because currently previous datasets are collected with manual labels, which can be a time-consuming and costly process, especially when large datasets are involved. By reducing the amount of human labor required, unsupervised gait recognition makes training gait recognition models with large datasets more feasible and cost-effective. Training with larger datasets can ultimately lead to more accurate and robust gait recognition models, which can have a wide range of applications in fields such as security, healthcare, and sports analysis. 

In a nutshell, for future work, more intelligent methods can be developed to identify sequences taken from front/back views and incorporate them with other views correctly. And more robust architecture can be developed to make the gait recognition method stable when adapting to real-world datasets.

%


%





\ifCLASSOPTIONcaptionsoff
  \newpage
\fi

\bibliographystyle{IEEEtran}
\bibliography{Transactions-Bibliography/IEEEabrv,Transactions-Bibliography/egbib}



%



